\documentclass[journal]{IEEEtran}

\IEEEoverridecommandlockouts
\usepackage{cite}
\usepackage{lipsum}
\usepackage{multirow}
\usepackage{amsmath,amssymb,amsfonts}
\usepackage{graphicx}
\usepackage{textcomp}
\usepackage{xcolor}
\usepackage{makecell}
\usepackage{array}
\usepackage{threeparttable}
\usepackage{algorithm}
\usepackage{algorithmic}
\usepackage{amsthm}
\usepackage{booktabs}
\usepackage{bm}
\usepackage{tabu}
\usepackage{url}

\usepackage{subfloat}

\usepackage{makecell}
\usepackage{threeparttable}

\usepackage{bbding}
\usepackage{rotating}
\usepackage{pifont}
\usepackage{changes}
\usepackage{colortbl}
\definecolor{mygreen}{rgb}{.51,.83,.65}
\definecolor{myblue}{rgb}{.50,.84,.97}
\definecolor{myyellow}{rgb}{.98,.788,.56}
\definecolor{mypink}{rgb}{.96,.58,.90}
\definecolor{myred}{rgb}{.93,.68,.70}
\definecolor{myblue2}{rgb}{.56,.64,.76}
\definecolor{myblue3}{rgb}{.41,.93,.95}
\definecolor{mygray}{rgb}{.75,.75,.75}

\newcommand{\thickhline}{%
    \noalign {\ifnum 0=`}\fi \hrule height 1pt
    \futurelet \reserved@a \@xhline
}

\usepackage{tikz,xcolor,hyperref}
 \usepackage[caption=false,font=footnotesize]{subfig}

\definecolor{lime}{HTML}{A6CE39}
\DeclareRobustCommand{\orcidicon}{%
    \begin{tikzpicture}
    \draw[lime, fill=lime] (0,0) 
    circle [radius=0.16] 
    node[white] {{\fontfamily{qag}\selectfont \tiny ID}};    \draw[white, fill=white] (-0.0625,0.095) 
    circle [radius=0.007];    \end{tikzpicture}
    \hspace{-2mm}}
\foreach \x in {A, ..., Z}{%
    \expandafter\xdef\csname orcid\x\endcsname{\noexpand\href{https://orcid.org/\csname orcidauthor\x\endcsname}{\noexpand\orcidicon}}
    }
    
\begin{document}
\title{BAND-2k: Banding Artifact Noticeable Database for Banding Detection and Quality Assessment}

\author{
        Zijian~Chen\orcidA{},~\IEEEmembership{Student Member,~IEEE}, Wei~Sun,~\IEEEmembership{Member,~IEEE}, Jun~Jia, Fangfang~Lu, Zicheng~Zhang, Jing~Liu, Ru~Huang, Xiongkuo~Min,~\IEEEmembership{Member,~IEEE}, and Guangtao~Zhai,~\IEEEmembership{Senior~Member,~IEEE}

\thanks{\fontdimen2\font=0.5ex Zijian Chen, Wei Sun, Jun Jia, Zicheng Zhang, Xiongkuo Min, and Guangtao Zhai are with the Institute of Image Communication and Information Processing, Shanghai Jiao Tong University, Shanghai 200240, China (e-mail: \{zijian.chen, sunguwei, jiajun0302, zzc1998, minxiongkuo, zhaiguangtao\}@sjtu.edu.cn).}
\thanks{\fontdimen2\font=0.65ex Fangfang Lu is with the College of Computer Science and Technology, Shanghai University of Electric Power, Shanghai 200290, China (e-mail:lufangfang@shiep.edu.cn).}
\thanks{\fontdimen2\font=0.65ex Jing Liu is with the School of Electrical and Information Engineering, Tianjin University, Tianjin, China. (e-mail:jliu\_tju@tju.edu.cn).}
\thanks{\fontdimen2\font=0.65ex Ru Huang is with the School of Information Science \& Engineering, East China University of Science and Technology, Shanghai 200237, China (e-mail:huangrabbit@ecust.edu.cn).}
}


\maketitle

\begin{abstract}
\fontdimen2\font=0.5ex
Banding, also known as staircase-like contours, frequently occurs in flat areas of images/videos processed by the compression or quantization algorithms. 
As undesirable artifacts, banding destroys the original image structure, thus inevitably degrading users' quality of experience (QoE).
In this paper, we systematically investigate the banding image quality assessment (IQA) problem, aiming to detect the image banding artifacts and evaluate their perceptual visual quality.
Considering that the existing image banding databases only contain limited content sources and banding generation methods, and lack perceptual quality labels (\textit{i.e.} mean opinion scores), we first build the largest banding IQA database so far, named \textit{\underline{B}anding \underline{A}rtifact \underline{N}oticeable \underline{D}atabase} (\textit{BAND-2k}), which consists of 2,000 banding images generated by 15 compression and quantization schemes. A total of 23 workers participated in the subjective IQA experiment, yielding over 214,000 patch-level banding class labels and 44,371 reliable image-level quality rating scores.
Subsequently, we develop an effective no-reference (NR) banding evaluator for banding detection and quality assessment by leveraging frequency characteristics of banding artifacts. To be more specific, a dual convolutional neural network (CNN) is employed to concurrently learn the feature representation from the high-frequency and low-frequency maps, thereby enhancing the ability to discern banding artifacts. The quality score of a banding image is generated by pooling the banding detection maps masked by the spatial frequency filters.
The experimental results demonstrate that our banding evaluator achieves a remarkably high accuracy in banding detection and also exhibits high SRCC and PLCC results with the perceptual quality labels, even without directly learning a regression model for banding quality evaluation. These findings unveil the strong correlations between the intensity of banding artifacts and the perceptual visual quality, thus validating the necessity of banding quality assessment.
The BAND-2k database and the proposed banding evaluator will be available at \textcolor{red}{\url{https://github.com/zijianchen98/BAND-2k}}.
\end{abstract}


\begin{IEEEkeywords}
\fontdimen2\font=0.6ex Image quality assessment, banding artifact, frequency maps, database, dual-branch, deep learning. 
\end{IEEEkeywords}

\IEEEpeerreviewmaketitle

\section{Introduction}
\fontdimen2\font=0.5ex
\IEEEPARstart{R}{ecent} years have witnessed a rapid emergence of media streaming services and social platforms. YouTube, Netflix, and TikTok account for more than half of the world's video traffic. Improving the quality of images under limited encoding, transmission bandwidth, and storage condition is a necessary prerequisite for meeting the quality of experience (QoE) of users. In the stages between image acquisition and display, an image may suffer from various types of degradation, while banding artifacts are a kind of false contour distortion that is quite perceptible to the human eye. 
Since the visual quality of image contents greatly affects the QoE of end-users, it is highly desirable to design an effective banding image quality assessment (IQA) method, which aims to automatically detect the traces of such false contours and predict the objective quality of banding images that can be used to develop pre-processing or post-processing \textit{debanding} algorithms and optimize the performance of streaming media application.

Normally, banding artifacts take on the appearance of annual rings, radiation circles, halos, or geographical contour lines and especially exist in the background regions (\textit{e.g.}, sky, water, and wall surface), where the color transition is not smooth enough. Nearly all existing image or video encoders, including H.264/AVC \cite{wiegand2003overview}, VP9 \cite{mukherjee2013latest}, and H.265/HEVC \cite{sullivan2012overview} can introduce such artifacts more or less.
Current banding IQA research can be divided into two categories: subjective quality assessment and objective quality assessment. The existing subjective banding IQA research \cite{wang2016perceptual,tu2020bband,tandon2021cambi,kapoor2021capturing} mainly investigates the limited banding scenarios with internal-used and undisclosed databases while lacking the quality label and may be insufficiently generalizable to large-scale commercial applications.
Meanwhile, general IQA methods aiming at common distortions are inapplicable for banding exacerbated images due to the essential differences between them. First, banding artifacts usually hold tiny, staircase-like, and regional structures, which can be regarded as a kind of high-frequency artifact in smooth areas, while general distortions occur obviously in the whole image and are globally uniform. Second, the perceptual severity of banding is quantified based on its fraction of coverage and intensity in an image, which is widely divergent from the design philosophy of many existing IQA approaches. As a result, it is challenging to design an effective banding IQA method.


\begin{table*}
\centering
\caption{Comparison of the Existing Banding Artifact Databases with BAND-2k}
\label{datasets}
\renewcommand\arraystretch{1.1}
\begin{threeparttable}
\begin{tabular}{lcccc}
\toprule[0.75pt]
\textsc{Database Attribute} & Wang \textit{et al.}  \cite{wang2016perceptual}&Tandon \textit{et al.} \cite{tandon2021cambi} &Kapoor \textit{et al.} \cite{kapoor2021capturing}&BAND-2k \\ 
\midrule[0.5pt]
Publication year&2016&2021&2021&2023 \\
Number of contents&7&9&600&873\\
Video sources&YouTube&Netflix catalogue&unknown&CG, UGC, PGC\\
Stimuli type&video&video&image&image\\
Resolution&1280$\times$720&4k&1920$\times$1080&1920$\times$1080\\
Distortion source&VP9&AV1, downsampling\tnote{a}&bit-depth\tnote{b}&H.264, H.265, VP9, bit-depth\tnote{b}\\
Distortion levels&3&9&6&3, 3, 3, 6\\
Total number of stimuli&21&86&1,440&2,000\\
Test environment&laboratory&remote&$-$&laboratory\\
Number of subjects&25&23&$-$&23\\
Number of ratings&$>$1,000&unknown&$-$&44,371\\
Rating scale&Continuous Rating 0-100&Continuous Rating 0-100&$-$&Continuous Rating 0-100\\
Patch-level label&\textcolor{red}{\ding{55}}&\textcolor{red}{\ding{55}}&\textcolor{green}{\checkmark}&\textcolor{green}{\checkmark}\\
Open-sourced&\textcolor{red}{\ding{55}}&\textcolor{red}{\ding{55}}&\textcolor{green}{\checkmark}&\textcolor{green}{\checkmark}\\
Study remarks&\multirow{1}{*}{\begin{tabular}{@{}p{3.1cm}@{}}Study did not include more other codecs, which is not general enough; Lack of content source and the analysis for statistics of MOS.\end{tabular}}&\multirow{1}{*}{\begin{tabular}{@{}p{3cm}@{}}Study of banding visibility tracking across dithering was introduced; Insufficient video source.\end{tabular}}&\multirow{1}{*}{\begin{tabular}{@{}p{3cm}@{}}Subjective quality assessment with MOS or DMOS was not provided.\end{tabular}}&\multirow{1}{*}{\begin{tabular}{@{}p{4cm}@{}}Content sources are collected from three typical video categories; A total of 15 degrees of distortion are introduced to generate banding artifacts; Both MOS and patch-level label are provided and open-sourced.\end{tabular}}\\
&&&&\\
&&&&\\
&&&&\\
&&&&\\
&&&&\\
\bottomrule[0.75pt]
\end{tabular}

\begin{tablenotes}
        \scriptsize
        \item[a] Source videos are downsampled to appropriate resolution (1080p, quad-HD or 4k) and bit-depth (8 bit).  
        \item[b] The bit-depth quantization is applied in luminance and chrominance channels with reduction and promotion operations. 
      \end{tablenotes}
    \end{threeparttable}
    
\end{table*}

To address these limitations, we first conduct a comprehensive subjective study of banding exacerbated images and create the largest banding IQA database to date with reliable mean opinion scores (MOS) and patch-level banding labels. We also propose a novel no-reference banding evaluator for banding detection and quality assessment by leveraging the frequency characteristics of banding artifacts. First, due to the dissimilar peculiarities of banding and smooth regions, the same distortion in different regions, \textit{e.g.}, textual and pictorial regions, may lead to different visual perception of human beings.
Considering that banding manifests as a high-frequency artifact that exists in the low-frequency smooth region, we propose a dual-branch CNN, which takes the high-frequency map and low-frequency map as inputs simultaneously, to hierarchically incorporate different visual features from the first layer and the last layer of our Resnet-50 backbone, thus making the model learn more effective banding feature representation and achieving more accurate banding region discrimination.
Furthermore, inspired by the previous studies \cite{li2008multifocus,kazemi2022multifocus}, spatial frequency extracts information consistent with the human visual system (HVS), which not only reflects the overall active level in an image but also intuitively quantifies the contrast information. In other words, the value of spatial frequency is large in smooth areas, while becoming small in areas with harsh contrast changes, \textit{i.e.}, banding areas. 
Based on this mechanism, we adopt a spatial frequency masking strategy to refine the detected banding map and then pool the masked banding detection map to obtain the image-level banding quality score. In summary, this paper makes the following contributions:


\begin{itemize}
	\item We construct so far the largest banding-affected database and name it the \textbf{B}anding \textbf{A}rtifact \textbf{N}oticeable \textbf{D}atabase (BAND-2k). It contains 2,000 distorted images sampled from over 870 source videos with four encoding schemes: H.264, H.265, VP9, and bit-depth manipulation. A total of 44,371 scores are collected by 23 no-experienced subjects in a well-controlled laboratory environment. Compared to existing banding artifact datasets (Table \ref{datasets}), BAND-2k is times larger than them \cite{wang2016perceptual, tandon2021cambi} in terms of the number of contents and includes more comprehensive compression means while providing patch-level banding labels for training deep learning models. 
	\item We propose a novel no-reference banding evaluator for banding detection and quality assessment based on frequency characteristics of banding artifacts, which utilizes a dual-branch CNN model to extract hierarchical banding-related feature representation from the high-frequency maps and low-frequency maps simultaneously. A spatial frequency masking strategy is introduced to refine the visibility of banding contours, and then combine with the detected banding map to generate subjectively consistent banding quality scores.
        \item Experimental results show that the proposed banding evaluator achieves the best performance in banding detection and significantly surpasses baselines in terms of SRCC and PLCC in the banding IQA task, which demonstrates the effectiveness of the proposed model. 
\end{itemize}


The remainder of this paper is organized as follows. Section \uppercase\expandafter{\romannumeral2} provides an overview of related works, including the state-of-the-art banding databases and detection methods. Section \uppercase\expandafter{\romannumeral3} introduces the construction of the BAND-2k database and the subjective assessment study. Section \uppercase\expandafter{\romannumeral4} proposes an effective no-reference banding evaluator for banding detection and quality assessment. Section \uppercase\expandafter{\romannumeral5} gives the experimental results and analysis. Section \uppercase\expandafter{\romannumeral6} concludes this paper.

\begin{figure*}[!t]
\centering
{\includegraphics[width=0.95\textwidth]{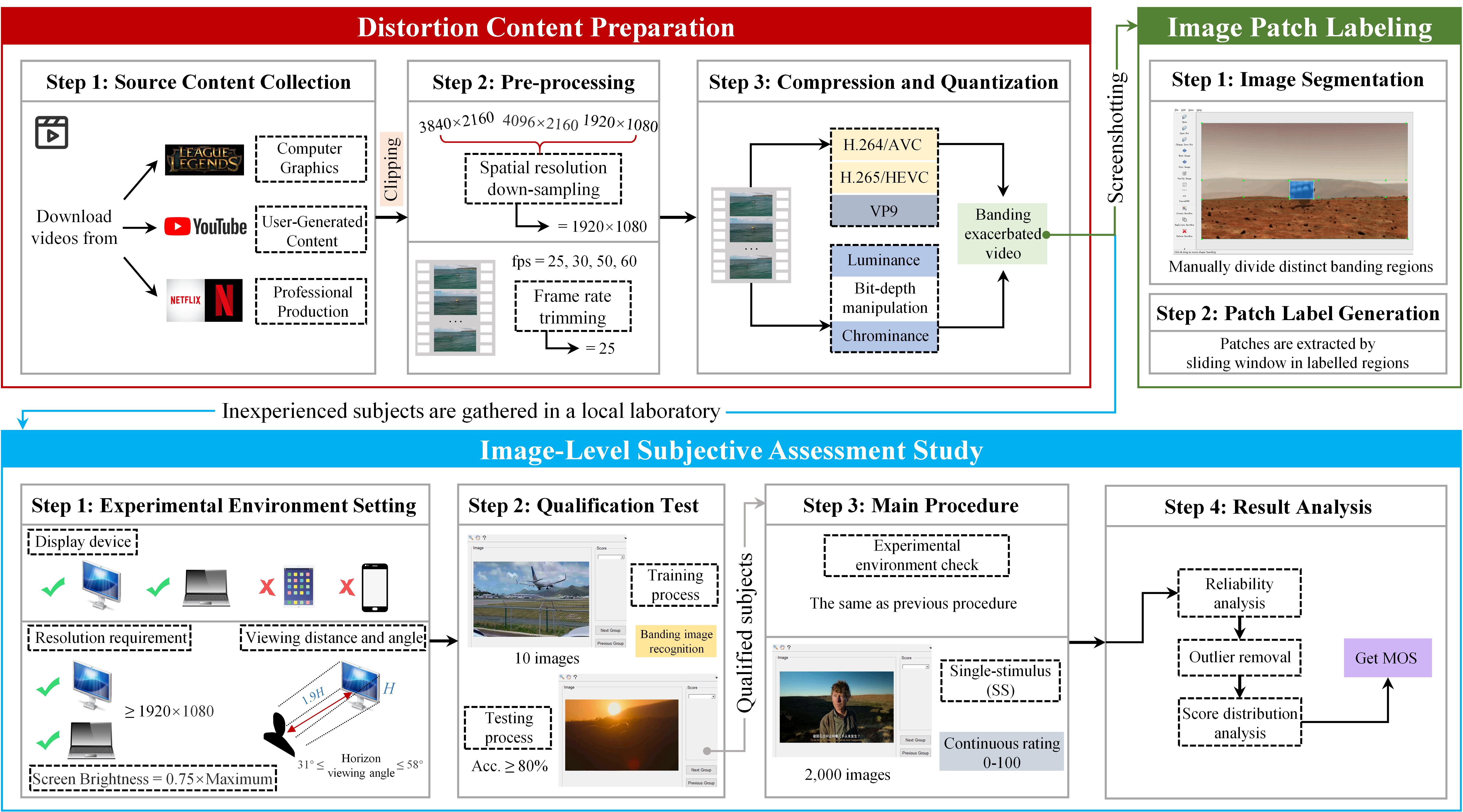}} 
\caption{Workflow of the banding database construction. It mainly includes three parts: 1) Distortion content preparation; 2) Image patch labeling; 3) Image-level subjective assessment study.} \label{subjective}
\end{figure*}

\section{Related Work}
In this section, we first provide an overview of the state-of-the-art banding-related databases (Table \ref{datasets}) and then review the banding detection and evaluation methods.
\subsection{Banding-Related Databases}
The first banding artifact-relevant VQA database was proposed by Wang \textit{et al.} \cite{wang2016perceptual}, which consists of 21 stimuli with different quantization grades generated by VP9 \cite{mukherjee2013latest} from 7 clips of 1280 $\times$ 720 30fps video. Authors in \cite{tandon2021cambi} investigated the effect of encoding parameters and dithering on the visibility of banding. Nine 4k-10bit source clips from the existing Netflix catalogue between 1 and 5 seconds were used to generate banding distorted videos. Each source content was downsampled to appropriate resolutions (1080p, 2k, or 4k) with certain bit-depth and further compressed by \textit{libaom} (an AV1 codec library) at QPs $\left\{ 12,20,32\right\}$. 
More recently, Kapoor \textit{et al.} \cite{kapoor2021capturing} constructed one of the first databases for data-driven image banding assessment models. This research included about 1,440 images shot from over 600 pristine HD videos with a resolution of 1920 $\times$ 1080. 
Six levels of bit-depth quantization in luminance and chrominance channels are introduced to obtain different intensities of banding. 
Meanwhile, the banding images were semi-automatically segmented and labeled into banded and non-banded to form a patch-level banding dataset, which allows for training machine learning-based and deep learning-based banding classification methods.
However, to the best of our knowledge, thus far there still lacking a benchmarking dataset in the banding detection and the corresponding banding IQA domain. Researchers either resort to image/video quality datasets that do not aim at banding distortion or build a small, attribute-restricted, in-house dataset by themselves. This motivates us to construct a large-scale subjective assessment database focus on the perceived banding-affected image quality.

\subsection{Banding Detection and Quality Assessment Methods}
 Early research on banding detection mainly focuses on false contour identification, which aims to find the wrong boundary rather than a ``true" region edge in the image. Authors in  \cite{daly2004decontouring, lee2006two, huang2016understanding} utilized monotonicity or non-monotonicity features of local support regions including the gradient, contrast, variance, and entropy information to measure the loss of low-amplitude detail caused by banding. However, these works ignored the perceptual characteristics of the human visual system (HVS) and thus did not perform a good correlation with subjective tests. Another banding detection strategy is conducted at the pixel-level estimation and segmentation. Bhagavathy \textit{et al.} \cite{bhagavathy2009multiscale} proposed to identify banding artifacts by calculating the likelihood of pixel difference. Baugh \textit{et al.} \cite{baugh2014advanced} measured the severity of banding based on the number of a group of connected pixels with the same color. Wang \textit{et al.} \cite{wang2016perceptual} first detected uniform segments to find possible banding areas and further incorporated edge features (e.g. length and contrast) to capture false boundaries. Nevertheless, these kinds of methods are typically sensitive to edge noise and are computationally expensive, causing limited application in real-time scenarios.

\begin{figure*}
	\centering
	\subfloat[Contrast]{\includegraphics[width=1.78in]{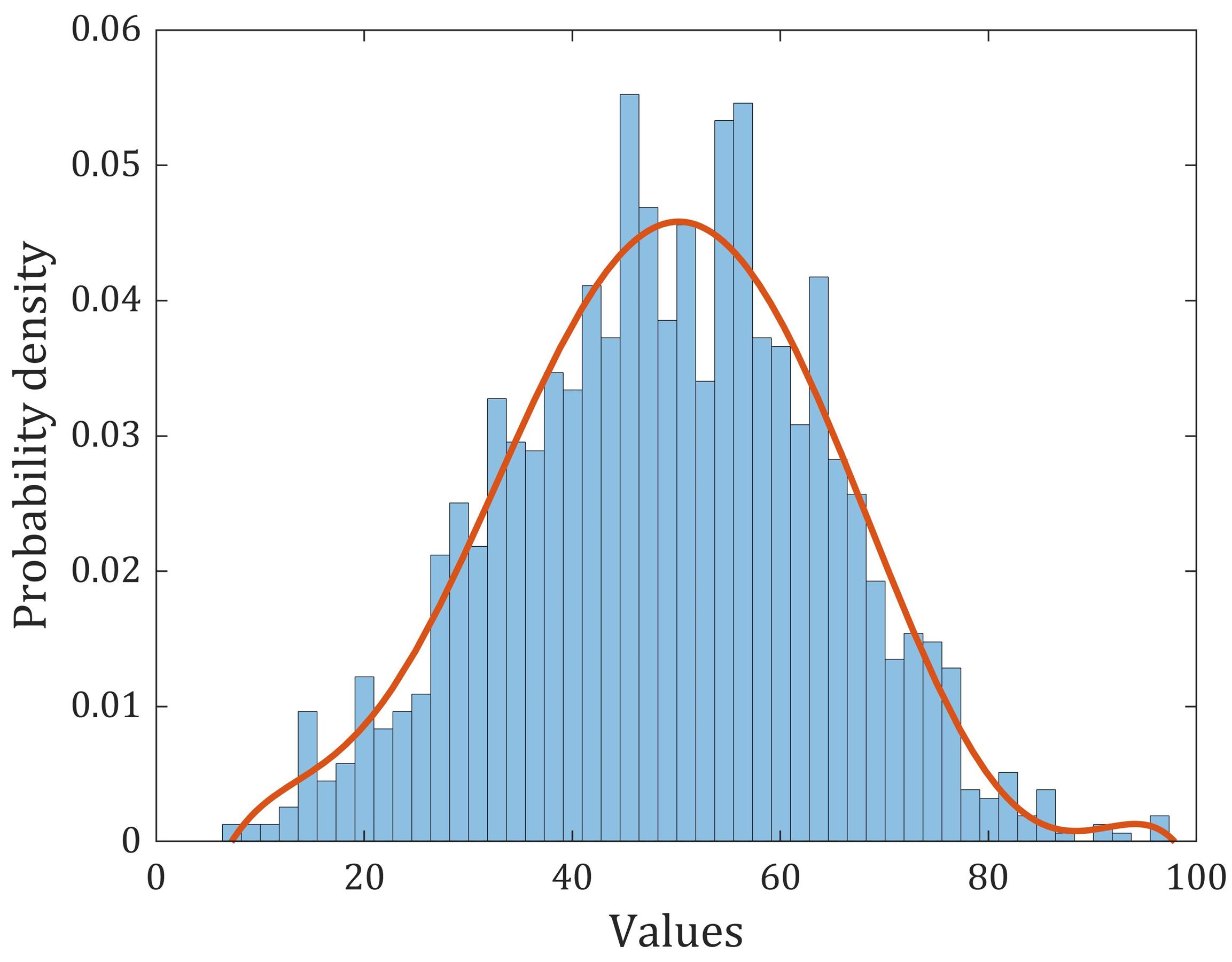}}
	\subfloat[Colorfulness]{\includegraphics[width=1.78in]{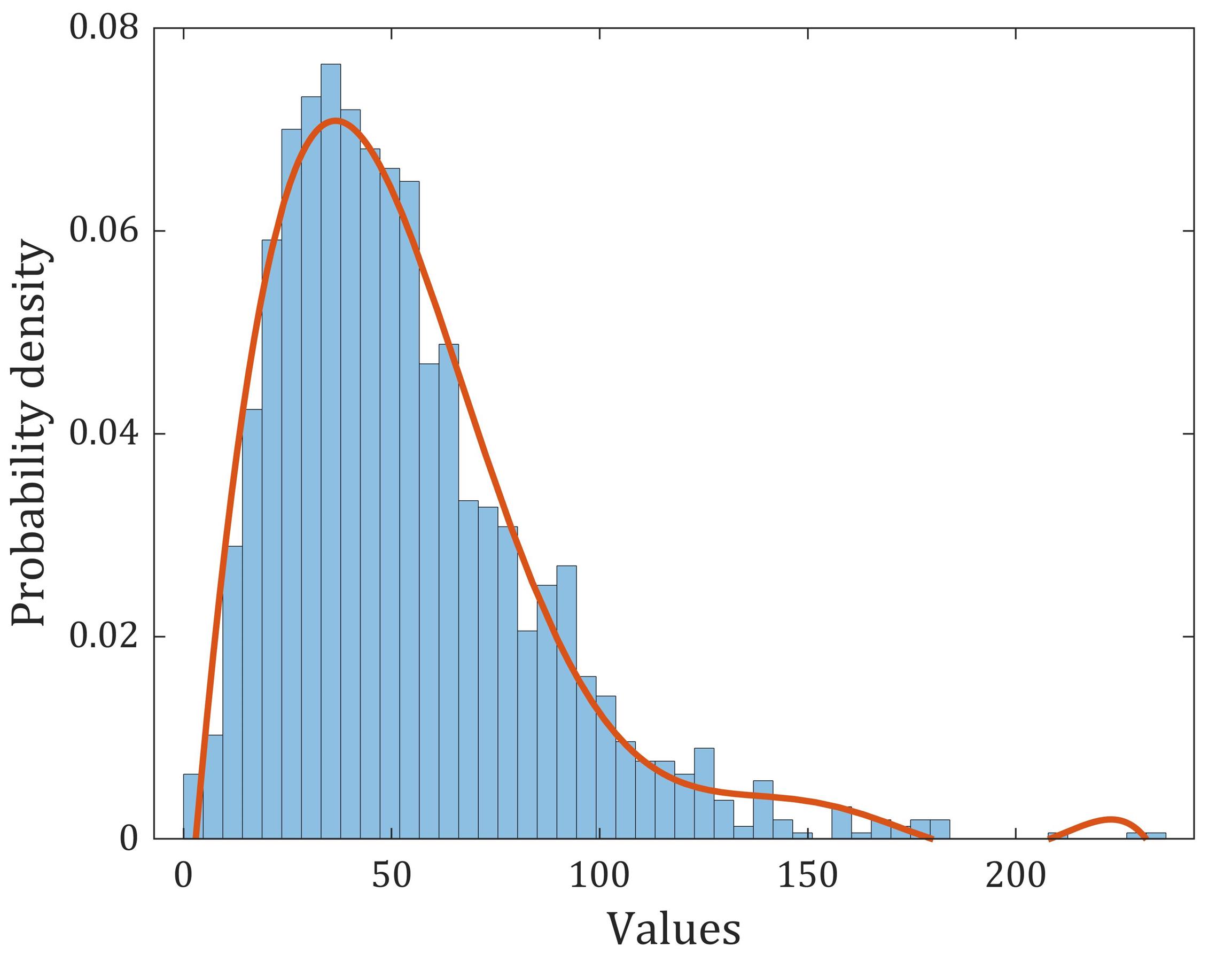}}
	\subfloat[Sharpness]{\includegraphics[width=1.78in]{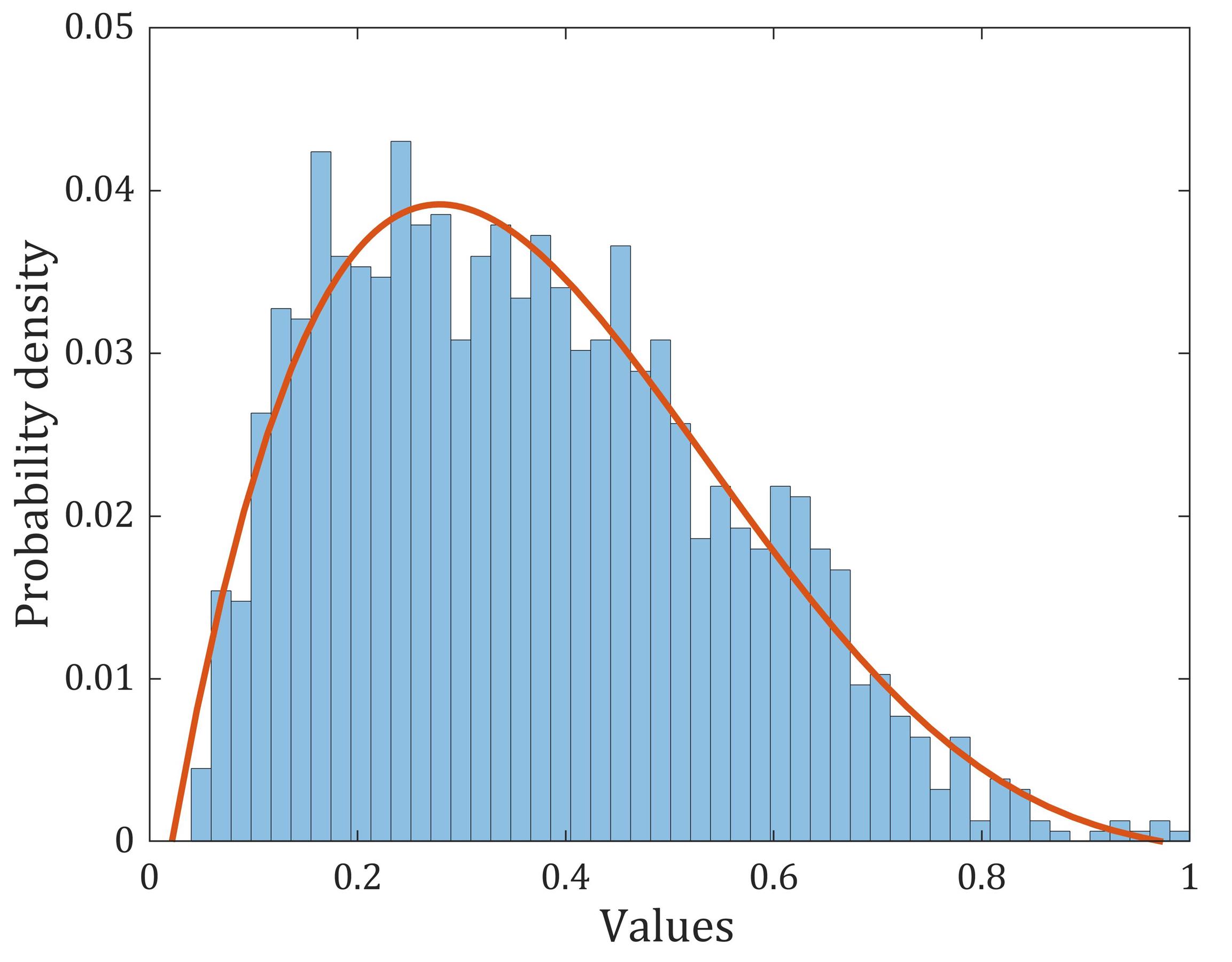}}
	\subfloat[Brightness]{\includegraphics[width=1.78in]{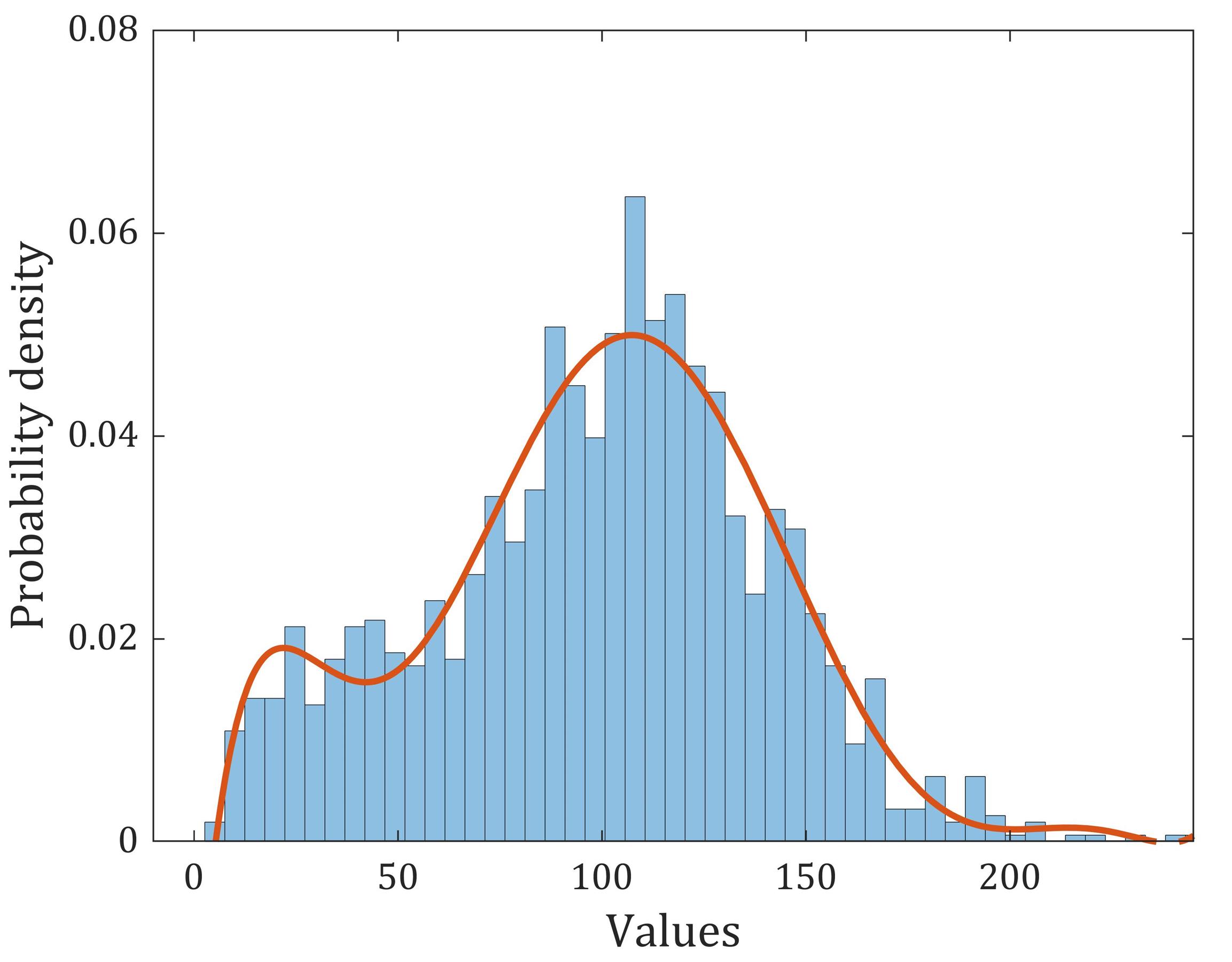}}
	\caption{Image attribute histograms and the fitted kernel distributions. (a)-(d) are the distribution of contrast, colorfulness, sharpness, and brightness, respectively.}
	\label{diversity}
\end{figure*}

\begin{figure*}[!t]
\centering
{\includegraphics[width=0.85\textwidth]{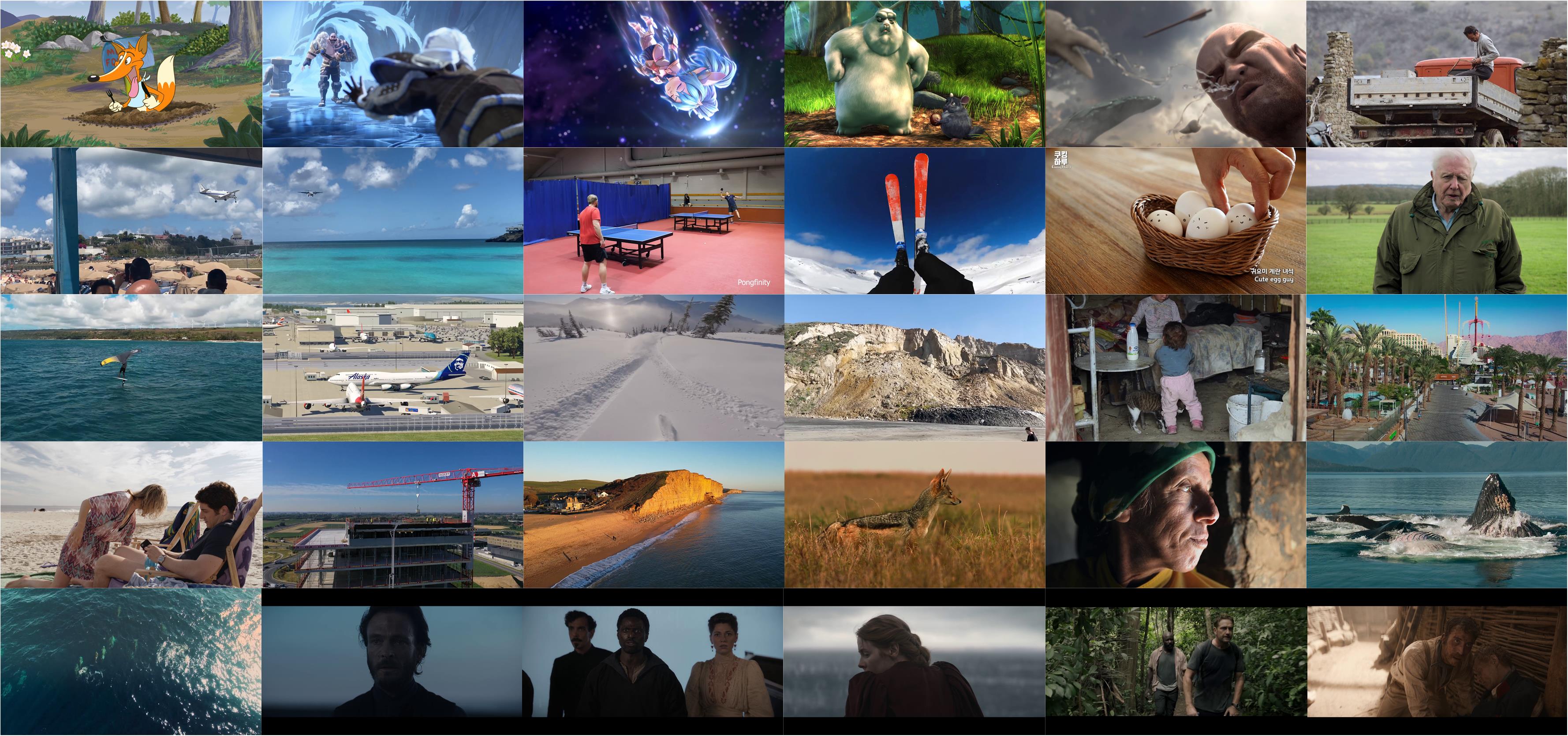}} 
\caption{Display of 30 representative thumbnails of video clips from the BAND-2k.} \label{example}
\end{figure*}

 Towards addressing these problems, Tu \textit{et al.} \cite{tu2020bband} presented a completely no-reference banding detection method, which combines various properties of HVS with a number of pre-processing steps to refine banding edge detection. Instead of regarding banding detection as a false edge detection problem, Tandon \textit{et al.} \cite{tandon2021cambi} heuristically utilized the effect of contrast sensitivity function (CSF) on banding visibility and its dependence on spatial frequency. Based on this, Krasula \textit{et al.} \cite{krasula2022banding} further compared the banding annoyance with more commonly studied compression artifacts and proposed a banding-aware video quality metric.
In recent years, deep learning approaches have prevailed in various VQA tasks. As the pioneering work, Kapoor \textit{et al.} \cite{kapoor2021capturing} developed an automated CNN-based banding detector for the first time, which is a simple two-stage algorithm and gives rise to devising other learning-based techniques. 

In this work, we build a large-scale banding database and propose a data-driven banding indicator that can generate pixel-wise banding visibility maps with corresponding subjectively consistent quality scores by combining human visual mechanisms and deep learning techniques.



\section{Banding Database Creation}
Subjective banding image quality assessment facilitates the development of automatic objective banding image and video quality models.
We created the largest banding database in existence, denoted as the BAND-2k database, which consists of 2,000 banding distorted images and over 214,000 patch-level banding class labels. Then, a subjective experiment was conducted to obtain the mean opinion scores (MOS) of the BAND-2k database.  
The workflow of the banding database construction is shown in Fig. \ref{subjective}.


\subsection{Source Content Collection}
To build a content-rich and balanced database, we manually collected source videos including computer graphics (CG), user-generated content (UGC), and professionally-generated content (PGC) from two popular media websites Bilibili.com and Youtube.com.  
Then, 885 clips with multiple spatial resolutions (i.e., 4096$\times$2160, 3840$\times$2160, 1920$\times$1080) and frame rates (i.e., 60, 50, 30, 25) are chosen as candidate.
Note that videos on the mentioned websites are firstly annotated by the community with assigned a number of favorites, views, and downloads. These statistics correlate with the content and quality of a video, which guides our choices to some extent. 
All videos selected on the website are released under an appropriate creative commons (CC) license that allows further editing and redistribution. 
After content selection, we further unified the format of all video clips, especially the spatial resolution and the pixel format, which avoid the effect of other facts on visual quality. Concretely, we first converted the frame rate of the original clips to 25fps, which is to reduce the storage pressure while ensuring the graphics quality. Considering the commonly used aspect ratio of the user interface and displays is 16:9, we cropped the partially unqualified videos rather than shrinking images unevenly. 
Then, we downsampled the trimmed spatial resolution 3840$\times$2160 to a lower resolution $-$ 1920$\times$1080 for the following subjective study.

\begin{figure*}
\centering
{\includegraphics[width=0.8\textwidth]{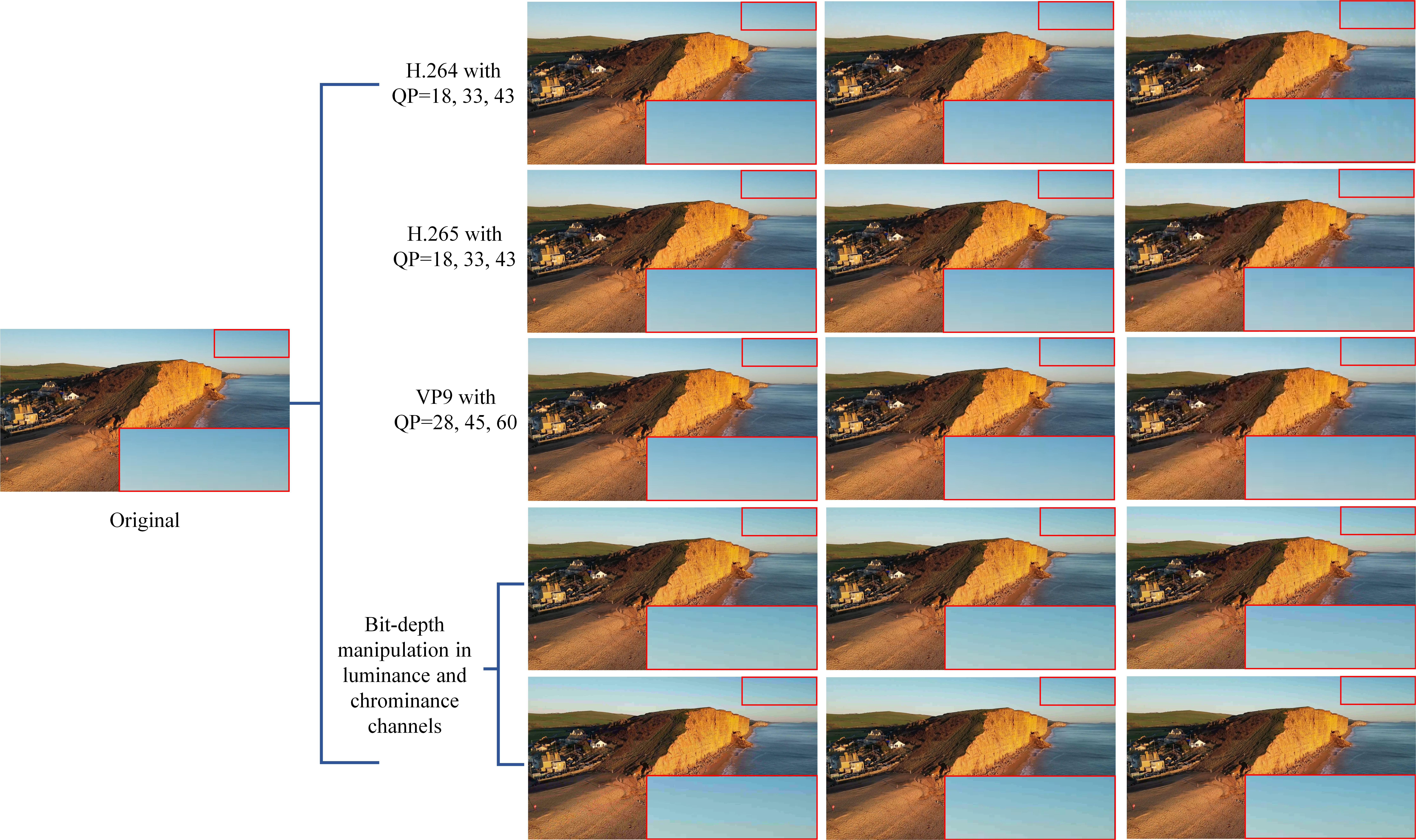}} 
\caption{Visual comparison of banding exacerbated images generated using H.264, H.265, VP9, and bit-depth quantization as compression means. Zoom-in for better viewing.} \label{quantized}
\end{figure*}

\subsection{Pre-processing and Patch Labeling}
Before conducting the following experiments, we manually  removed the videos that are either too dark or bright, overly blurry or colorful, which helps to obtain more reliable subjective assessments.
To avoid redundancy and to make sure the diversity of selected contents, we also conducted attribute analysis studies on the selected images. Four metrics that correlated with human perception, i.e., contrast, brightness, sharpness, and colorfulness, are adopted as content diversity metrics. All video attributes are calculated on every 10 frames to reduce computational complexity, which are then averaged over frames of each video sequence.
Fig. \ref{diversity} shows the distribution of attributes extracted from the selected videos.

\begin{itemize}
	\item \textit{Contrast:} The contrast metric is simply defined as the standard deviation of pixel gray-scale intensities \cite{hasler2003measuring}.
	\item \textit{Colorfulness:} The colorfulness metric is measured by the R, G, and B components \cite{hasler2003measuring}. We first compute two matrices of differences between channels $rg=R-G$ and $yb=\frac{1}{2} \left( R+G\right)  -B$. Then, the colorfulness metric can be calculated as $\sqrt{\mu^{2}_{rg} +\mu^{2}_{yb} } +\sqrt{\sigma^{2}_{rg} +\sigma^{2}_{yb} } $, where $\mu$ and $\sigma$ are the mean and standard deviation of their corresponding terms, respectively.
	\item \textit{Sharpness:} The cumulative probability of blur detection (CPBD) metric \cite{narvekar2011no} is used to measure the image sharpness, which estimates the probability of detecting blur at each edge.
	\item \textit{Brightness:} The brightness of an image is obtained directly from the pixel gray intensities in R, G, and B channels.
\end{itemize}
Finally, the number of source videos are reduced to 873. Fig. \ref{example} displays thumbnails for 30 selected representative video clips.

To simulate authentic banding artifacts that exist in real viewing scenarios, we introduced four encoding/transcoding strategies including H.264/AVC \cite{wiegand2003overview}, H.265/HEVC \cite{sullivan2012overview}, VP9 \cite{mukherjee2013latest}, and bit-depth manipulation \cite{kapoor2021capturing} with fifteen quantization schemes in total. For  H.264/AVC and H.265/HEVC, considering the range of their quantization parameter (QP in ffmpeg), we selected three typical QP values, namely, QP = $\left\{18, 33, 43\right\}$. This is because that coded video clips with a QP value smaller than 18 normally provide perceptual lossless quality, while coded video clips with a QP value larger than 43 will not be able to offer adequate quality, which may affect the subjective test of banding artifacts. Similarly, for the VP9 encoder, the QP values under our close inspection are chosen as 
$\left\{28, 45, 60\right\}$. Moreover, we applied the same quantization strategies as in \cite{kapoor2021capturing}, which introduces banding distortion by scaling bit-depth in luminance and chrominance channels. Here, the color coding scheme YCbCr4:2:0 is applied for maximum compatibility.
To sum up, we generated fifteen levels of banding with different intensities and shapes to enhance the diversity of the database. The banding database is then built by extracting frames from the distorted video clips, resulting in 2,000 images with a resolution of 1920$\times$1080. Fig. \ref{quantized} shows the visualization results of banding exacerbated images.

In the process of image patch labeling, we initially intend to extract image patches of size 235$\times$235 from banding images directly and perform annotation operations. However, considering the theoretical number of patches, it will cost a lot of manpower and time to label patch by patch.
Therefore, we first segmented the banding images roughly and label them into banded and non-banded regions. Then, labelled image patches are generated from these segmented and labelled images by a sliding window. Specifically, we followed the same demarcation of the banded and non-banded images in \cite{kapoor2021capturing} that a patch is labelled as banded if it has more than 30\% overlap with banded regions in the image.

Eventually, a banding dataset containing 2,000 distorted images with 1920$\times$1080 resolution and 214,324 labelled image patches of size 235$\times$235 is built. To the best of our knowledge, it is the largest banding dataset in existence, which enables training various machine/deep learning based banding detection models and facilitates the development of image/video debanding techniques.
Table \ref{patch} reports the composition of labelled image patch dataset. It can be observed that the number of banded patches is a bit smaller than non-banded patches due to the fact that banding usually appears in smooth background areas.

\begin{table}
\centering
\caption{Composition of Labelled Image Patch Dataset}
\label{patch}
\renewcommand\arraystretch{1.25}
\begin{tabular}{|c|c|c|}
\hline
                  & Banded          & Non-Banded       \\ \hline
Training (80\%)   & 69,472           & 101,988           \\ 
Validation (10\%)    & 8,684            & 12,748            \\ 
Testing (10\%)    & 8,684           & 12,748            \\ \hline
Total             & 86,840 (40.52\%) & 127,484 (59.48\%) \\ \hline
\end{tabular}
\end{table}

\subsection{Subjective Quality Assessment}
As shown in Fig. \ref{subjective}, the subjective quality study contains four steps. In addition to preparing the experimental environment, subjects should pass the qualification test first to participate in the study. After the subjective rating, all resulting scores need to be analyzed and examined before generating the final mean opinion score (MOS).
\subsubsection{Experimental Environment Setting}
In this study, a total of 25 inexperienced subjects are gathered in a laboratory environment, where relevant experimental configuration must satisfy the following requirements:
\begin{itemize}
	\item Considering the viewing effect, desktops and laptops are allowed as displays.
	\item The resolution of displays must be larger than or equal to 1920$\times$1080 to show the images without spatial downsampling.
	\item The viewing distance and optimal horizontal viewing angle are set as 1.9 times the height of the display and $\left[ 31^{\circ },58^{\circ }\right]  $, respectively. Other settings such as the ambient brightness, lighting, and background are configured according to the ITU-R BT.500 recommendation \cite{bt2002methodology}.
\end{itemize}
As a result, we used a 27-inch AOC Q27U2D monitor with a resolution of 2560$\times$1440 for assessment with 25 subjects. Due to the large number of images to be assessed, we divided the dataset into ten sessions to avoid visual fatigue. Each session of tests took nearly 2 hours with a 30-minute break for each participant.


\subsubsection{Qualification Test}
Before starting the main experiment, subjects are required to pass a quiz to get the qualification of conducting follow-up experiments. 
Firstly, we manually selected 10 labeled banding exacerbated images beyond the database as training images to familiarize subjects with the operation interface and the goal of this subjective test. 
The quiz consists of two parts including banding classification and image-level quality rating. In banding classification, subjects were told to divide the test image into banded or non-banded. In image-level quality rating, subjects were instructed to focus on the coverage and intensity of banding areas, as well as the overall quality of images to get the final quality score. The rating scale is continuous from 0 to 100 while a higher value indicates more severe banding (i.e., more visible or occupying larger portion of the image). 
To make the quiz objective and fair, we take the banding images labeled by domain experts as the ground truth, while two existing banding metrics BBAND \cite{tu2020bband} and DBI \cite{kapoor2021capturing} are used to determine the normal range of the rating scores. That is the quality scores rated by subjects should not exceed 20\% of the above banding metrics. 
As a result, only 23 subjects with an accuracy above 80\% in banding classification and subjective scores in the normal range were allowed to pass the quiz.
Note that the steps for taking a quiz are the same as the main experiment, which aims to guarantee the consistency of results.
\subsubsection{Formal Study}
We adopted the single-stimulus (SS) method in this test. Ten \textit{golden images} that have the acknowledged high quality or poor quality (assessed by BBAND \cite{tu2020bband} and DBI \cite{kapoor2021capturing}) were added to each session for controlling the scoring deviations. Besides, 3 repeated images are randomly inserted into each session to ensure consistency of scores before and after subjects scoring.
At last, 23 qualified subjects were asked to provide their opinions on the shuffled image groups. The resulting scores were collected and packed for further analysis.

\subsubsection{Result Analysis}
In total, 46,000 scores were collected by 23 qualified subjects in the main study phase. However, considering the qualification quiz cannot completely disallow those unreliable workers to muddle through to the main study while reliable subjects may also occasionally score odd values, which may be caused by the inter-individual differences in perceiving the quality of the unique characteristics of different contents.
Thus, we further investigated the confidence of rating scores and removed outliers following the Grubbs' test \cite{grubbs1950sample,wang2017videoset}.

\begin{figure}[!t]
\centering
{\includegraphics[width=0.4\textwidth]{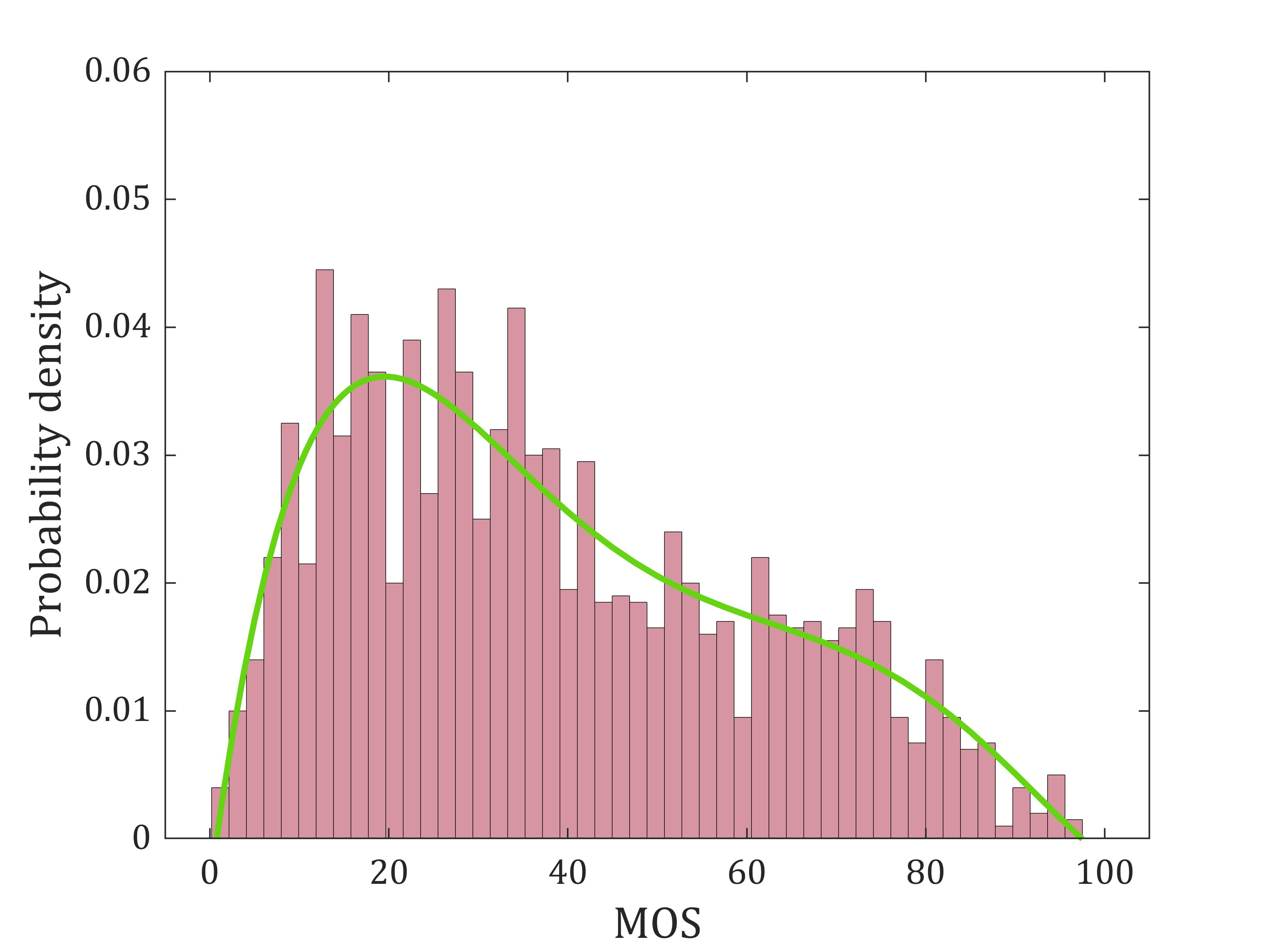}} 
\caption{MOSs histogram and the fitted kernel distribution of the BAND-2k database.} \label{mos}
\end{figure}

\begin{figure*}[!t]
\centering
{\includegraphics[width=0.85\textwidth]{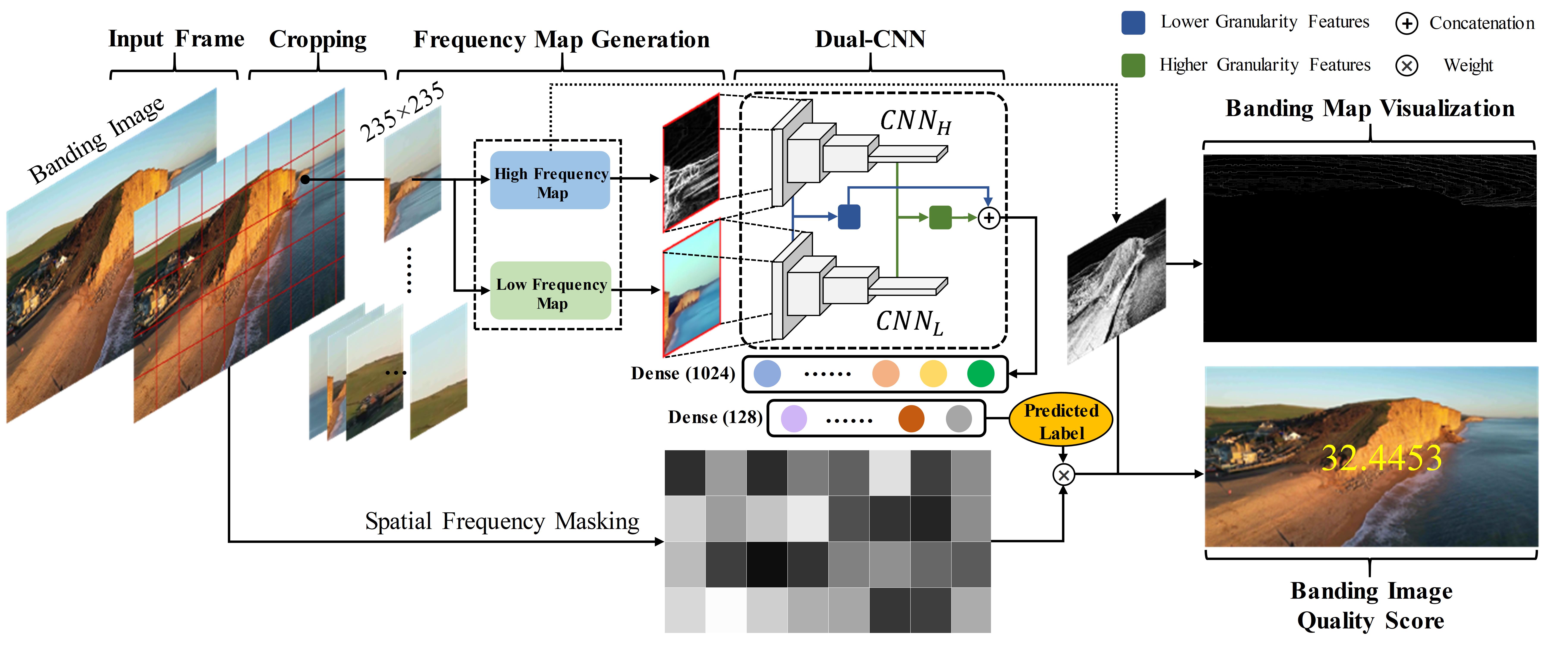}} 
\caption{The overall architecture of the proposed method. Given a banding distorted image, it is first divided into patches. Then, the patch-level high-frequency map (HFM) and low-frequency map (LFM) are generated by Sobel operation and piece-wise smooth algorithm \cite{bar2006semi}, respectively. After that, a dual-branch CNN ($\textbf{CNN}_\textbf{H}$ and $\textbf{CNN}_\textbf{L}$) is deployed to extract hierarchical features with different visual information and thus classify the patches into banded or non-banded. Lastly, a spatial frequency masking strategy is introduced to refine the banding map and calculate the image-level banding quality score. Note that the dual-branch networks do not share parameters.} \label{framework}
\end{figure*}

Concretely, let $\mathbf{s} =\left( s_{1},s_{2},\dots ,s_{N}\right)$ denote a set of raw scores collected for one distorted image. The test statistics is defined as the maximum absolute deviation of a sample standard deviation (SD) unit from the sample mean, which is mathematically expressed as
\begin{equation}
	G=\frac{\max \limits_{i=1,\dots,N} \left| s_{i}-\bar{\mathbf{s} } \right| }{\sigma_{s} } ,
\end{equation}
where $\bar{\mathbf{s} } $ and $\sigma_{s}$ denote the mean and standard deviation of the sample, respectively. Then, given a significant level $\alpha$, a sample is detected as an outlier if 
\begin{equation}
	G>\frac{N-1}{\sqrt{N} } \sqrt{\frac{t^{2}_{\alpha /\left( 2N\right)  ,N-2}}{N-2+t^{2}_{\alpha /\left( 2N\right)  ,N-2}} } ,
\end{equation}
where $t^{2}_{\alpha /\left( 2N\right)  ,N-2}$ represents the upper critical value of the $t$-distribution with $N-2$ degrees of freedom and a significance level of $\alpha/(2N)$.
Empirically, we set the significance level $\alpha$ at 0.05. Then, a sample is identified as an outlier if its distance to the sample mean is larger than 2.5 times SD and is removed. Following the aforementioned steps, the total number of scores was reduced to 44,371, and MOS was created by averaging the scores for each image. 
Fig. \ref{mos} presents the histogram of MOSs over the entire database, showing a broad MOS distribution of banding images.

\section{The Proposed Method}
In this section, we describe the architecture of the proposed banding evaluator in detail, as shown in Fig. \ref{framework}.


\subsection{Frequency Map Generation}
As stated before, banding usually appear as high-frequency information in the smooth background, while humans perceive high-frequency texture regions and low-frequency plateau regions through different neural channels concurrently, and transfer the upper visual features into the cerebral cortex for final processing \cite{deyoe1988concurrent,zhang2022dual}. Inspired by this, we employ high-frequency maps (HFM) and low-frequency maps (LFM) as the deep learning network inputs, which represent the texture and structural information of the image respectively, to mimic the recognition mechanism of the human brain for better banding identification.
\textbf{High-frequency Maps.} Since gradient has been widely used to represent edge information and has been confirmed beneficial to acquire high-frequency components with low computational cost \cite{tang2018full,gu2017evaluating,al2017multi}, we apply the isotropic Sobel operator to each patch for enhancing the details of banding artifacts. Given an input patch $\mathcal{I}$, the high-frequency map is calculated by
\begin{equation}
\mathcal{H}=\sqrt{\left( \mathcal{I} \ast \mathcal{S}_{x} \right)^{2}  +\left( \mathcal{I} \ast \mathcal{S}_{y} \right)^{2}  } ,
\end{equation}
where $\mathcal{S}_{x}$ and $\mathcal{S}_{y}$ are the horizontal and vertical isotropic Sobel operators, respectively. ``$\ast$" denotes the convolution operation.

\begin{figure}[!t]
	\centering
	\subfloat[Banding Image]{\includegraphics[width=0.155\textwidth]{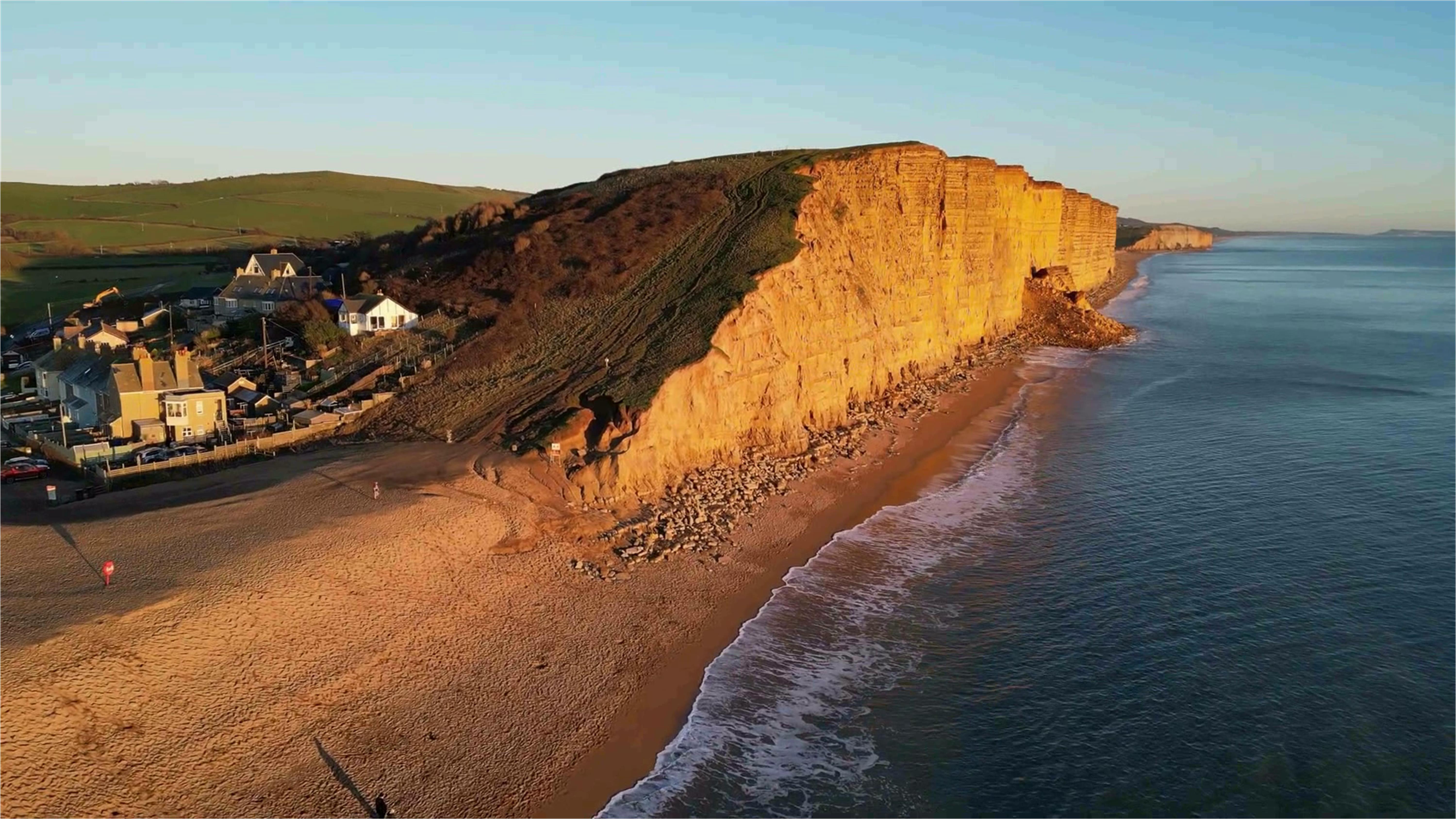}}
		\hspace{0.5mm}
		\subfloat[HFM]{\includegraphics[width=0.155\textwidth]{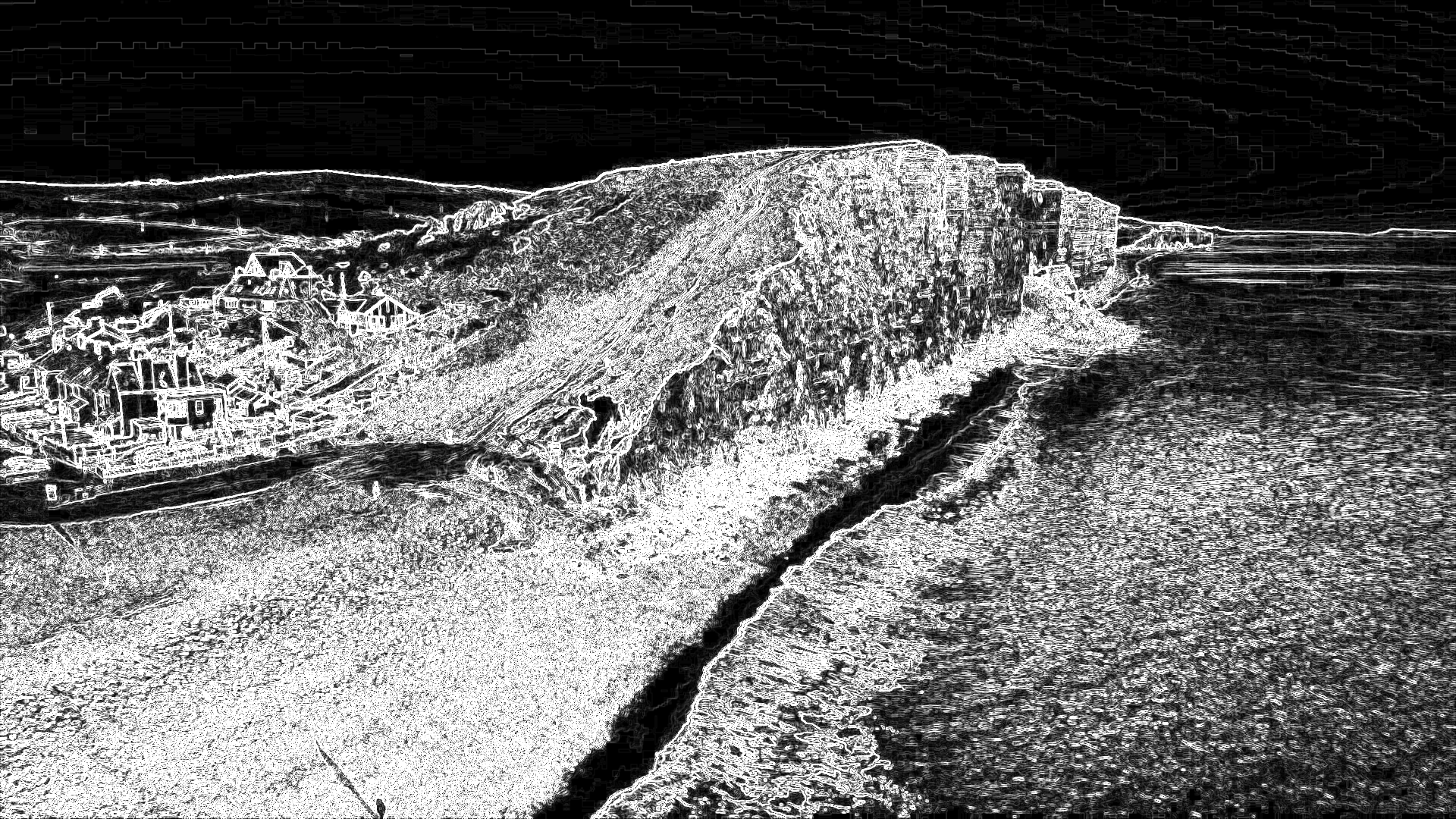}}
		\hspace{0.5mm}
	\subfloat[LFM]{\includegraphics[width=0.155\textwidth]{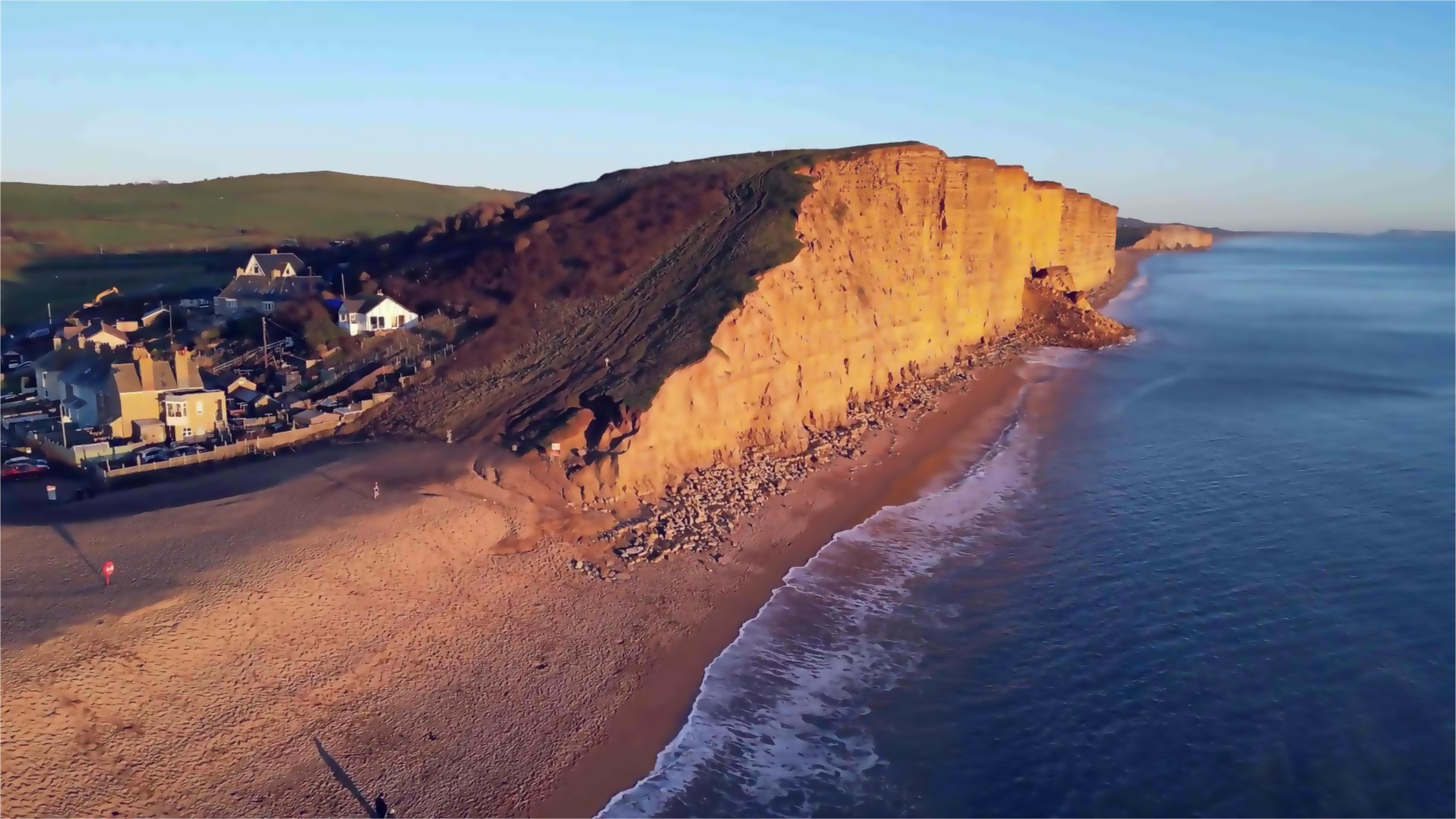}}
	\caption{Comparison of the banding image and its corresponding HFM and LFM from the BAND-2k database.}
	\label{frequency-map}
\end{figure}

\textbf{Low-frequency Maps.} To maintain the principal content of the image and filter out the influence of high-frequency information, we use the piece-wise smooth algorithm \cite{bar2006semi} to generate the low-frequency map by minimizing a function for image approximation recovery:
\begin{equation}
	\mathcal{F}=\frac{1}{2} \int_{\Omega } (\mathcal{I} -\mathcal{L})^{2}dP+\alpha \int_{\Omega \backslash E} |\nabla \mathcal{L} |^{2}dP+\beta \int_{E} d\sigma,
\end{equation}
where $\mathcal{L}$ represents the low-frequency map, $\Omega$ and $E$ denotes the image domain and edge set, respectively. $P$ indicates the pixel and $\int_{E} d\sigma$ represents the total edge length. The coefficients $\alpha$ and $\beta$ are positive regularization constants. An example of frequency maps is shown in Fig. \ref{frequency-map}.

\subsection{Dual-CNN Model}
To obtain the overall banding score, we first divide the banding image into 235$\times$235 patches and apply CNN-based classification to each patch, resulting in a banding classification label for each patch, \textit{i.e.}, banded or non-banded.
As shown in Fig. \ref{framework}, the proposed network consists of two parallel branches, namely $\textbf{CNN}_\textbf{H}$ and $\textbf{CNN}_\textbf{L}$, which take the patch-level high-frequency map and low-frequency map as input, respectively.  
For each branch, we propose to use Resnet-50 \cite{he2016deep} as the backbone. 
Specifically, we incorporate the feature maps extracted from the first convolutional layer and the last layer of Resnet-50 as hierarchical visual features, which represent different visual information \cite{zeiler2014visualizing,ranjan2017hyperface} and can be used as predictive information to enhance the discrimination ability of the network for banding and non-banded regions.
Afterward, the features extracted from two branches are concatenated first and reshaped into 128-dimensional vectors through two fully-connected layers, which is further followed with the sigmoid activation function to output the final predicted label, namely banded or non-banded. Of note is that sharing parameters is extremely unfavorable for extracting low- and high-frequency features simultaneously, we thereby deploy two branches that work independently and do not share parameters. 
The loss function adopted here is binary cross entropy.



\subsection{Banding Detection and Evaluation}
With the trained $\textbf{CNN}_\textbf{H}$ and $\textbf{CNN}_\textbf{L}$, each input patch is predicted to obtain a banding or non-banded label. To better guide the pre-processing and post-processing \textit{debanding} algorithms, it is necessary to generate a quality score for the entire banding image. Since the visibility of edge is also affected by content, we further consider the effect of spatially varying content information on the local quality of human perception. As a consequence, we introduce the spatial frequency masking strategy to determine the weighting factor for the detected banding regions in each patch adaptively and thus obtain the image-level banding severity score while refining the visibility of banding artifacts.


\subsubsection{Spatial Frequency Masking}
The spatial frequency is defined as the activity level of an image, which establishes a filter-bank based on the visual stimulus and is in accordance with HVS \cite{li2008multifocus}. In this paper, we propose to apply spatial frequency as an effective contrast criterion to banding measurement. Specifically, given an image of size $I_W \times I_H$, divided into $N \times N$ patches, where $I_W$ and $I_H$ denote the number of columns and rows respectively. The column ($CF_k$) and row ($RF_k$) frequencies of the image patches are given by
\begin{align}
	CF_{k}&=\sqrt{\frac{1}{N^2} \sum^{N}_{x=2} \sum^{N}_{y=1} \left( I\left( x,y\right)  -I\left( x-1,y\right)  \right)^{2}  } ,\\
	RF_{k}&=\sqrt{\frac{1}{N^2} \sum^{N}_{x=1} \sum^{N}_{y=2} \left( I\left( x,y\right)  -I\left( x,y-1\right)  \right)^{2}  } ,
\end{align}
where $I(x,y)$ is the pixel value of the image patch. Then, the resulting spatial frequency of an  $N\times N$ patch is computed as 
\begin{equation}
SF_{k}=\sqrt{CF^{2}_{k}+RF^{2}_{k}} ,
\end{equation}
where $k$ is the number of patches ($1\leq k\leq \frac{I_{W}I_{H}}{N^{2}}$). Since most banding regions are likely to have large contrast including edges and textures, which should be assigned greater weights than the smooth and blurred areas. Following Kazemi \textit{et al.} \cite{kazemi2022multifocus}, we set a threshold value to distinguish these regions, which is defined as the average spatial frequency of image patches:
\begin{equation}
	\epsilon =\frac{N^{2}}{I_{W}I_{H}} \sum^{I_{W}I_{H}/N^{2}}_{k=1} SF_{k}.
\end{equation}
Accordingly, we design a banding visibility transfer function to express spatial frequency masking as a function of the local textural feature. The final spatial frequency masking weight is calculated at each patch as
\begin{equation}
	w_{k}=\begin{cases}1&|SF_{k}|\leq \epsilon \\ 1+\left( |SF_{k}|-\epsilon \right)^{\gamma }/N &|SF_{k}|>\epsilon, \end{cases}
\end{equation}
where $\gamma$ is the scaling constant factor chosen to tune the shape of the transfer function. We used $\gamma=1.5$ in our implementation.

\subsubsection{Building a Banding Metric}
The visibility of banding artifacts depends on the combination of multiple visual mechanisms.
In this paper, we propose a simple but effective product model for attribute integration at each predicted banding patch to obtain the entire banding map (BM):
\begin{equation}
		\mathrm{BM}_{k}(i,j) =w_{k}\cdot \widehat{P_{k}} \cdot |\mathrm{HFM}_{k} \left( i,j\right)  |,
\end{equation}
where $\widehat{P_{k}}$ denotes the predicted label of $k$-th patch and $w_k$ is the weight parameter that scales the visibility of measured contours, \textit{i.e.}, gradient magnitude of the high-frequency map, $|\mathrm{HFM}_{k} \left( i,j\right)|$ at region $(i,j)$. Furthermore, inspired by previous psychovisual findings that the QoE of observers is dominated by those regions having poor quality \cite{ghadiyaram2017no,tu2020bband}, we thereby leverage the worst $p\%$ percentile visual pooling to calculate an average banding score from the generated BM, where $p$ is set to $80$ in this experiment. As a result, the perceptual score of the overall banding image is defined as
\begin{equation}
	\mathcal{Q} \left( \mathcal{I} \right)  =\frac{1}{M} \cdot \frac{1}{|\mathcal{T}_{p\% } |} \sum^{M}_{k=1} \sum\nolimits_{\left( i,j\right)  \in \mathcal{T}_{p\% } } \mathrm{BM}_{k} \left( i,j\right) , 
\end{equation}
where $M$ is the total number of patches in image $\mathcal{I}$. $\mathcal{T}_{p\% }$ denotes the index set of the top $p\%$ non-zero pixel-wise value contained in $k$-th patch of the BM.

\section{Experiments}
In this section, we first present the experimental protocol in detail and then evaluate the performance of the proposed method on two tasks, namely patch-level banding classification and banding image quality assessment. After that, the ablation study and cross-database validation are conducted to prove the robustness and effectiveness of the proposed method. Finally, we test the computational efficiency of our method.

\begin{table*}
\centering
\caption{Experimental Results of Patch-Level Banding Classification on the Database from Kapoor \textit{et al.} \cite{kapoor2021capturing} and the BAND-2k Database. Accuracy and Speed are Reported in the Form of Maximum Testing Accuracy and Execution Time in Seconds per Image Patch. The Best Result is Highlighted}
\label{classification}
\renewcommand\arraystretch{0.9}
\begin{tabular}{c|c|l|cccc|cccc}
\toprule[0.75pt]
\multirow{2}{*}{\textsc{Ref}}&\multirow{2}{*}{\textsc{Type}}&\multirow{2}{*}{\textsc{Metric}$\setminus$\textsc{Model}}&\multicolumn{4}{c|}{\textbf{Kapoor \textit{et al.} \cite{kapoor2021capturing} }}&\multicolumn{4}{c}{\textbf{BAND-2k}} \\ 
&&&AUROC$\uparrow$ &AUPRC$\uparrow $ &Accuracy$\uparrow $ &Speed$\downarrow $ &AUROC$\uparrow $ &AUPRC$\uparrow $ &Accuracy$\uparrow $ &Speed$\downarrow $ \\
\midrule[0.5pt]
\multirow{4}{*}{FR}&\multirow{4}{*}{General}&PSNR&0.0585&0.0543&25.34\%&\textbf{0.0071}&0.0573&0.0558&25.21\%&\textbf{0.0074}\\
&&SSIM \cite{wang2004image}&0.2421&0.2417&48.63\%&0.0091&0.2388&0.2357&47.61\%&0.0112\\
&&MS-SSIM \cite{wang2003multiscale}&0.2543&0.2674&52.63\%&0.0153&0.2521&0.2597&47.88\%&0.0183\\
&&LPIPS \cite{zhang2018unreasonable}&0.6571&0.6428&71.33\%&0.0098&0.6466&0.6410&70.89\%&0.0099\\
\midrule[0.5pt]
FR&Banding&VMAF\textsubscript{BA} \cite{krasula2022banding} &0.2955&0.2746&47.11\%&0.0173&0.3268&0.3441&48.15\%&0.0174\\
\midrule[0.5pt]
\multirow{6}{*}{NR}&\multirow{6}{*}{General}&BRISQUE \cite{mittal2012no}&0.2638&0.3163&56.31\%&0.0295&0.2587&0.2743&51.66\%&0.0282\\
&&NIQE  \cite{mittal2012making}&0.1627&0.2134&44.15\%&0.0412&0.1607&0.1928&43.59\%&0.0416\\
&&NIMA \cite{talebi2018nima}&0.2853&0.2767&46.32\%&0.0125&0.2739&0.2655&46.11\%&0.0137\\
&&DBCNN \cite{zhang2018blind}&0.7435&0.7327&74.92\%&0.0147&0.7518&0.7387&75.36\%&0.0149\\
&&HyperIQA \cite{su2020blindly}&0.7652&0.7626&78.66\%&0.1758&0.7681&0.7673&79.64\%&0.1754\\
&&StairIQA \cite{sun2023blind}&0.7117&0.6933&67.34\%&0.1053&0.7236&0.7157&67.75\%&0.1058\\
\midrule[0.5pt]
\multirow{4}{*}{NR}&\multirow{4}{*}{Banding}&BBAND \cite{tu2020bband}&0.3322&0.3175&45.72\%&0.1002&0.3409&0.3257&46.78\%&0.1022\\
&&CAMBI \cite{tandon2021cambi}&0.1553&0.1468&28.63\%&0.0087&0.1614&0.1502&28.68\%&0.0088\\
&&DBI \cite{kapoor2021capturing}&0.9442&0.9461&91.23\%&0.0231 &0.9023&0.8958&88.53\%&0.0232\\
&&Ours&\textbf{0.9872}&\textbf{0.9833}&\textbf{96.41\%}&0.0252&\textbf{0.9527}&\textbf{0.9544}&\textbf{94.18\%}&0.0278\\
\bottomrule[0.75pt]
\end{tabular}
\end{table*}

\subsection{Experimental Protocol}
\subsubsection{Databases and Settings}
We choose two databases to train and test the effectiveness of the proposed banding IQA method, which are the database released in \cite{kapoor2021capturing} and our proposed BAND-2k database. 
The detail information of these two datasets can be found in Table \ref{datasets}.
The proposed model is implemented by PyTorch \cite{paszke2017automatic}. Before training, we randomly split the training, validation, and testing set into 8:1:1 (as shown in Table \ref{patch}). We use the Adam optimizer with the initial learning rate set as 1e-4 and set the batch size as 32. The training process is stopped after 25 epochs. The resolution of each cropped patch is fixed to 235$\times$235. All experiments on both the \cite{kapoor2021capturing} database and the BAND-2k database are conducted repeatedly 10 times to obtain the mean performance.

\subsubsection{Baseline Algorithms}
We include a number of representative IQA algorithms in our evaluation as references to be compared against. These baseline methods include:
\begin{itemize}
	\item General FR IQA methods: We choose PSNR, SSIM \cite{wang2004image}, MS-SSIM \cite{wang2003multiscale}, LPIPS \cite{zhang2018unreasonable} as baselines. These are the most commonly used FR IQA metrics in practical applications such as video coding, image enhancement, etc.
	\item General NR IQA methods: BRISQUE \cite{mittal2012no}, NIQE \cite{mittal2012making}, NIMA \cite{talebi2018nima}, DBCNN \cite{zhang2018blind}, HyberIQA \cite{su2020blindly}, and StairIQA \cite{sun2023blind}. These are general-purpose NR IQA methods that are not limited by distortion types.
	\item Banding IQA methods: Considering that there exists few research on banding detection and quality assessment, we barely select the BBAND \cite{tu2020bband}, CAMBI \cite{tandon2021cambi}, VMAF\textsubscript{BA} \cite{krasula2022banding} and DBI \cite{kapoor2021capturing} metrics as comparisons. Among them, only VMAF\textsubscript{BA} is FR method, while the rest of them are NR.
\end{itemize}

\subsubsection{Evaluation Criteria}
To evaluate the IQA methods comprehensively, a total of seven evaluation indexes in two categories are adopted. For patch-level banding classification, we follow the common procedures as in \cite{kapoor2021capturing} and utilize the area under the receiver operating characteristics (AUROC), the area under the precision-recall curve (AUPRC), and accuracy as the classification performance metrics. For banding image quality assessment, four mainstream metrics are selected as the evaluation criteria: Spearman rank-order correlation coefficient (SRCC) and Kendall rank-order correlation coefficient (KRCC) measure the prediction monotonicity, while Pearson linear correlation coefficient (PLCC) and root mean square error (RMSE) are calculated to assess prediction consistency. Considering the potential non-linear mapping characteristics between the objective scores and the subjective scores, we perform score alignment by mapping the predicted value using the five-parameter logistic function before calculating PLCC and RMSE values \cite{seshadrinathan2010study}.



\begin{table*}
\centering
\caption{Performance Comparison on the BAND-2k Database for FR and NR IQA. The \textbf{Boldfaced} and \uline{Underlined} Entries Indicate the Best and Second-Best Values for Each Row, Respectively}
\label{iqa}
\begin{threeparttable}
\begin{tabular}{c|ccccc|cccccccccc}
\toprule[0.75pt]
\multirow{2}{*}{\scriptsize \textsc{Criteria}}&\multicolumn{5}{c|}{\scriptsize \textbf{Full Reference}}&\multicolumn{10}{c}{\scriptsize \textbf{No Reference}} \\ 
 &\tiny PSNR&\tiny SSIM&\tiny MS-SSIM&\tiny LPIPS&\tiny VMAF\textsubscript{BA}&\tiny BRISQUE&\tiny NIQE&\tiny NIMA&\tiny DBCNN&\tiny HyberIQA&\tiny StairIQA&\tiny BBAND\tnote{$\dag $}&\tiny CAMBI&\tiny DBI&\tiny Ours\\
\midrule[0.5pt]
\scriptsize SRCC&\scriptsize 0.0773&\scriptsize 0.2573&\scriptsize 0.5002&\scriptsize 0.5742&\scriptsize 0.3123&\scriptsize 0.2711&\scriptsize 0.3056&\scriptsize 0.3143&\scriptsize 0.7384&\scriptsize 0.7325&\scriptsize 0.6172&\scriptsize 0.2384&\scriptsize 0.1804&\scriptsize \uline{0.7432}&\scriptsize \textbf{0.8775}\\
\scriptsize KRCC&\scriptsize 0.0516&\scriptsize 0.1887&\scriptsize 0.3569&\scriptsize 0.4079&\scriptsize 0.2171&\scriptsize 0.1903&\scriptsize 0.2132&\scriptsize 0.2175&\scriptsize 0.5546&\scriptsize 0.5494&\scriptsize 0.4462&\scriptsize 0.1642&\scriptsize 0.1267&\scriptsize \uline{0.5654}&\scriptsize \textbf{0.7062}\\
\scriptsize PLCC&\scriptsize 0.1162&\scriptsize 0.0987&\scriptsize 0.2587&\scriptsize 0.5565&\scriptsize 0.2241&\scriptsize 0.2566&\scriptsize 0.2588&\scriptsize 0.3069&\scriptsize 0.7401&\scriptsize 0.7303&\scriptsize 0.6234&\scriptsize 0.1875&\scriptsize 0.0369&\scriptsize \uline{0.7446}&\scriptsize \textbf{0.8787}\\
\scriptsize RMSE&\scriptsize 4.8356&\scriptsize 6.6081&\scriptsize 6.6126&\scriptsize 6.6685&\scriptsize 7.4975&\scriptsize 5.3878&\scriptsize 6.2761&\scriptsize 6.4859&\scriptsize \textbf{4.4945}&\scriptsize \uline{4.6409}&\scriptsize 6.8817&\scriptsize 6.6383&\scriptsize 6.5747&\scriptsize 6.6403&\scriptsize 6.6358\\
\bottomrule[0.75pt]
\end{tabular}

\begin{tablenotes}
        \scriptsize
        \item[$\dag $] Considering the input in this experiment is images rather than videos, we use the proposed frame-level BBAND index $\mathrm{Q}_{\mathrm{B} \mathrm{B} \mathrm{A} \mathrm{N} \mathrm{D} } \left( \mathcal{I} \right)$ \cite{tu2020bband}. 
      \end{tablenotes}
    \end{threeparttable}

\end{table*}

\begin{figure*}[!t]
\centering
{\includegraphics[width=0.75\textwidth]{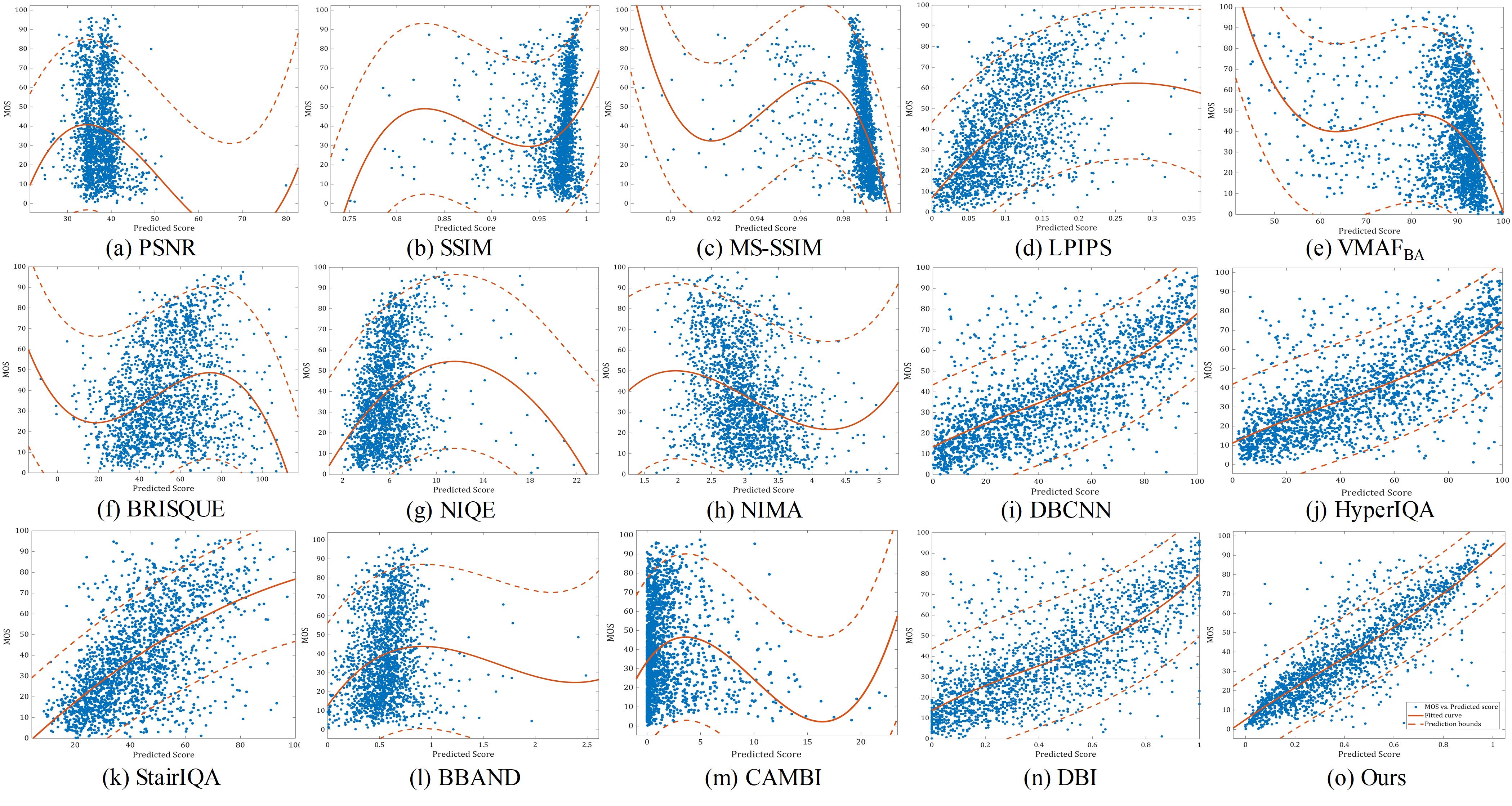}} 
\caption{Scatter plots and polynomial fitted curves of IQA methods versus MOS on the BAND-2k database. The dotted lines represent the 95\% prediction interval. (a) PSNR, (b) SSIM \cite{wang2004image}, (c) MS-SSIM \cite{wang2003multiscale}, (d) LPIPS \cite{zhang2018unreasonable}, (e) VMAF\textsubscript{BA} \cite{krasula2022banding}, (f) BRISQUE \cite{mittal2012no}, (g) NIQE \cite{mittal2012making}, (h) NIMA \cite{talebi2018nima}, (i) DBCNN \cite{zhang2018blind}, (j) HyperIQA \cite{su2020blindly}, (k) StairIQA \cite{sun2023blind}, (l) BBAND \cite{tu2020bband}, (m) CAMBI \cite{tandon2021cambi}, (n) DBI \cite{kapoor2021capturing}, and (o) Ours.} \label{scatter}
\end{figure*}


\subsection{Performance on Patch-Level Banding Classification}
Since our goal is to develop an effective banding IQA approach, we regard the identification of banding patches as an important preceding process to achieve accurate banding quality prediction. However, there exist few methods that are designed for banding classification and nearly all IQA methods produce scalar values only while failing in classifying banding regions directly. Therefore, we adopt a thresholding step to convert the single quality value into binary classification results as \cite{kapoor2021capturing} does. Concretely, a half-interval search algorithm \cite{bentley1975multidimensional} is employed to find the optimal threshold value that can generate the best classification result. 

Based on the above premise, Table \ref{classification} reports the experimental results on both the database from \cite{kapoor2021capturing} and the BAND-2k database. We highlight the best results in \textbf{boldface}. As compared to other state-of-the-art IQA methods, our proposed method yields the best overall performance in terms of AUROC, AUPRC, and accuracy. It is shown that most general FR IQA and NR IQA models perform poorly on the patch-level banding classification task while performing fairly well on other IQA tasks \cite{zhang2018blind,su2020blindly,sun2023blind}, indicating that the current approaches are not sensitive to banding distortion. Benefiting from the powerful feature extraction ability of CNNs, our proposed method and the customized NR IQA models for banding artifacts detection (DBI \cite{kapoor2021capturing}) reach a significant performance in the discrimination of false contours.
However, the performance of banding IQA method BBAND, CAMBI, and VMAF\textsubscript{BA} is surprisingly poor compared with other methods, which shows their vulnerability in identifying local banding artifacts from texture regions and are not suitable for patch-level banding identification. 
In addition, we investigate the computational complexity in terms of execution time per image patch. It can be observed that except for those traditional FR IQA models, our method achieves comparable speed in patch-level banding classification, which determines the prediction efficiency of the subsequent image-level quality assessment, making it a favorable choice in time-constrained scenarios.

\begin{figure*}[!t]
\centering
{\includegraphics[width=0.98\textwidth]{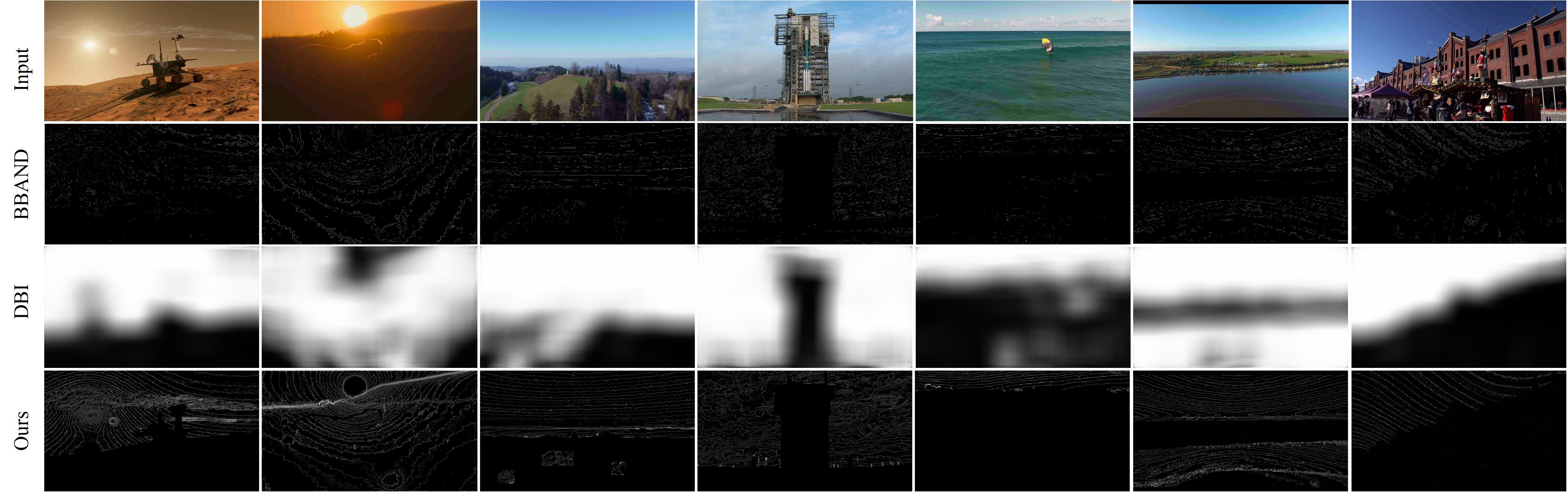}} 
\caption{Visual comparisons of the banding map results. From top to bottom are banding images and their corresponding banding maps generated by BBAND \cite{tu2020bband}, DBI \cite{kapoor2021capturing}, and our proposed method, respectively. The first five columns of images from left to right are from BAND-2k, while the rest images are from \cite{kapoor2021capturing}.}
\label{visualization}
\end{figure*}

\begin{figure}[!t]
\centering
{\includegraphics[width=0.48\textwidth]{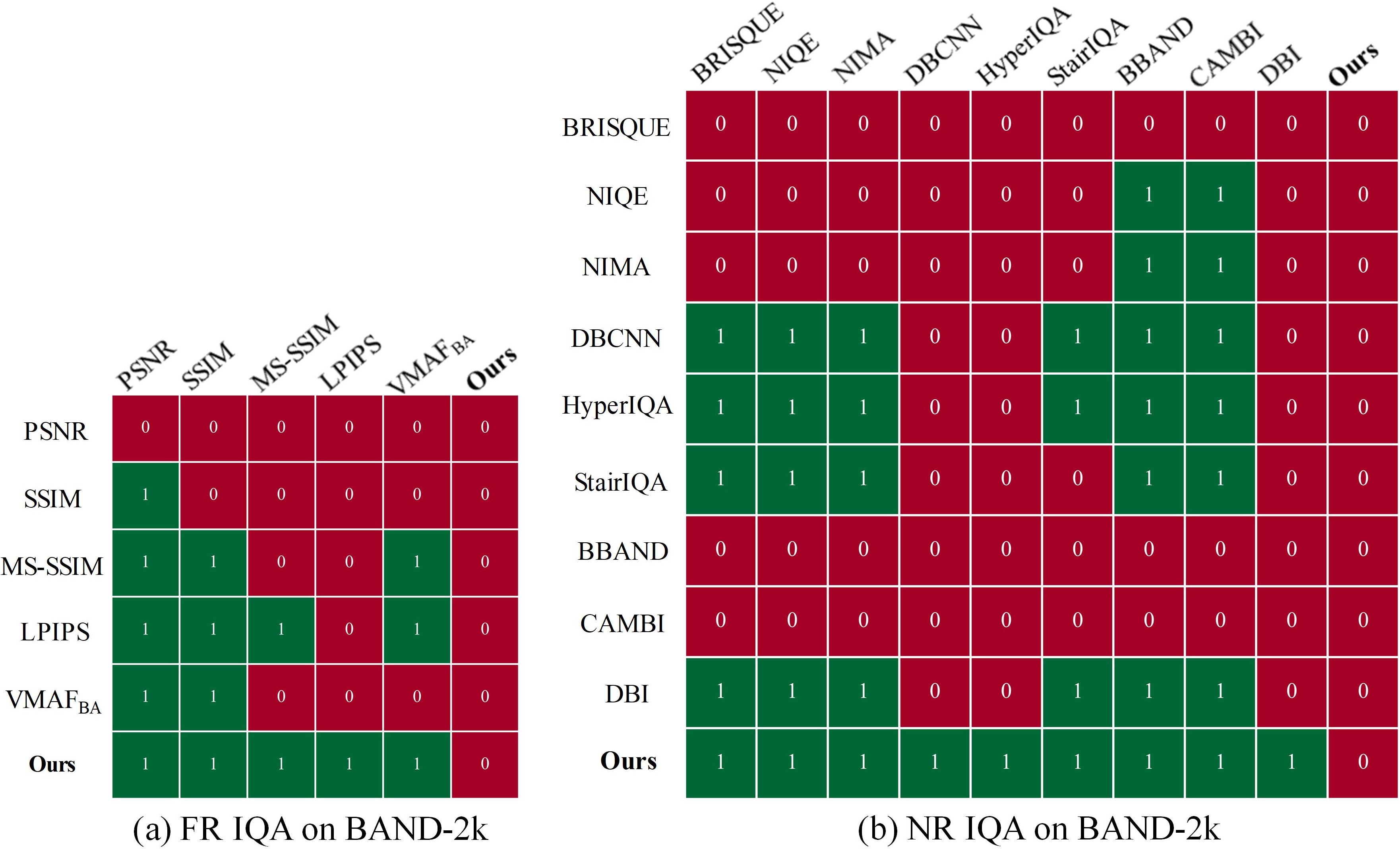}} 
\caption{The results of significance tests on the BAND-2k database for FR IQA and NR IQA methods.}
\label{statistic}
\end{figure}

\subsection{Comparison with State-of-the-Art IQA Methods}
Considering that there exist no image banding databases with subjective scores attached in the public domain, we merely compare the performance of the proposed method with the baseline approaches on the proposed BAND-2k database. 
The results are shown in Table \ref{iqa}, from which we can observe that our proposed method largely surpasses all baselines in terms of SRCC, KRCC, and PLCC except for the RMSE. Compared to the second-best model, our method achieves 18.07\% SRCC improvements, 24.91\% KRCC improvements, and 18.01\% PLCC improvements on the BAND-2k database.
We also present scatter plots of predictions versus MOS for better visualization in Fig. \ref{scatter}. Overall, the performance of traditional FR IQA models such as PSNR, SSIM, and MS-SSIM is remarkably inferior in banding images assessment and is uncorrelated with the MOS, which is consistent with the numerical results presented in Table \ref{iqa}. The reason is that PSNR and SSIM-based methods do perform not well on tiny, regional, and content-independent banding distortion while lacking the consideration for the mechanism of HVS.

It is also worth mentioning that the deep CNN architecture-based methods (DBCNN \cite{zhang2018blind}, HyperIQA \cite{su2020blindly}, and StairIQA \cite{sun2023blind}), despite performing well on LIVEC \cite{ghadiyaram2015massive}, KonIQ-10k \cite{hosu2020koniq}, and other universal image quality evaluation databases \cite{ciancio2010no,ponomarenko2015image}, underperformed our proposed model by a notable margin on the BAND-2k database.
Here we make two arguments to try to explain the observations above: (1) banding exacerbated image quality is intrinsically correlated with the coverage of banding contour. (2) the perception of banding artifacts is explicitly \textit{intensity-aware}. These are the issues that the CNN-based approaches above do not take into account. To some extent, banding distortion is more like a kind of local distortion than common global distortion such as Gaussian noise, blur, and dither. This suggests that it is potentially valuable to integrate some local texture, edge, contrast, or other visibility-related features into quality prediction models when assessing banding images. Fortunately, with the help of the proposed scoring strategy, our method gains the ability to convert the area range of identified banding regions to indicative annoying scores.

Surprisingly, the banding IQA methods BBAND \cite{tu2020bband}, CAMBI \cite{tandon2021cambi}, and VMAF\textsubscript{BA} \cite{krasula2022banding}, however, did not perform very well on the BAND-2k dataset. We infer that this is due to differences in the test environment. First, the source videos selected in CAMBI and VMAF\textsubscript{BA} are from the internal Netflix 4K catalogue while the source contents in BAND-2k are collected from the public streaming websites with different resolutions, leading to an uneven quality level. Second, the methods of artificially introducing banding distortion are different. In the databases used by CAMBI and VMAF\textsubscript{BA}, only AV1 and H.264 compression techniques are used to generate the banding artifacts, limiting the adaptive ability of the algorithm in other scenarios.
For further investigation, we compare our method with other two methods, i.e., BBAND \cite{tu2020bband}, DBI \cite{kapoor2021capturing}, and visually study the generated banding maps to verify their effectiveness in banding detection. The visualization results are shown in Fig. \ref{visualization}, where seven representative banding images are selected for reference. 
As shown, the banding maps generated from BBAND are mostly disordered and have a lot of discontinuity while the banding maps generated by DBI are too vague to locate the exact location of banding artifacts, making it difficult to develop pixel-level \textit{debanding} techniques.
Overall, we can compendiously conclude that: (1) in comparison with BBAND and DBI, banding maps computed by our proposed method could provide an accurate, clear indication for banding artifacts; (2) in comparison with BBAND and DBI, the quality prediction of banding images guided by our method could achieve a high consistency with HVS. 


Moreover, to make a statistically meaningful comparison among different IQA methods, we further conduct the widely used F-test \cite{ni2018gabor,yang2021full} to assess the statistical significance of the proposed method. Based on the assumption that the model's prediction residuals follow the Gaussian distribution, the left-tailed F-test with a confidence level of 95\% is performed on the residuals of every two IQA models. The results of significance tests on the BAND-2k database are shown in Fig. \ref{statistic}. A value of `1' (colored in green) indicates that the model in the row is significantly better than the model in the column, while a value of `0' (colored in red) indicates that the model in the row is not significantly better than the model in the column. It is shown that our proposed method performs significantly better than other models, which is consistent with the observations from the above comparison experiments.

\subsection{Ablation Study}
In this section, we explore the effectiveness of our model's design philosophy. To verify the importance of the dual-branch (DB) architecture, the baseline variants use the same backbone as the proposed method, except that only a single branch (SB) is reserved. Then, we use the original banding images as the input while removing the frequency map generation module (\textbf{SB-I}). Besides, the high-frequency maps and low-frequency maps generation modules are retained respectively (\textbf{SB-HFM} and \textbf{SB-LFM}).
It can be observed from Table \ref{ablation} that \textbf{SB-LFM} achieved the worst results, since the low-frequency map filters out the high-frequency banding information to a certain extent, which reduces the ability of the model to identify false contours. \textbf{SB-I} performs significantly better than \textbf{SB-HFM} and \textbf{SB-LFM}, resulting from that it contains richer image information, but it is still inferior to our method, which demonstrates the effectiveness of the dual-branch scheme.

To investigate the effect of the banding feature extraction, we further design two variants with different input combinations of frequency maps. First, the high-frequency map was taken as the input of both the $CNN_H$ and $CNN_L$ layers (\textbf{DB-HFM}). Then, we replace the inputs with the low-frequency maps (\textbf{DB-LFM}). 
As shown in Table \ref{ablation}, the performance of \textbf{DB-HFM} and \textbf{DB-LFM} is far apart from our approach, which matches our hypothesis that the high-frequency texture information contained in HFM and the low-frequency background information contained in LFM are crucial to enhance the capacity of discernment for banding artifacts.
Therefore, we may conclude that our model is the most suitable model among those compared variants in terms of both banding classification and IQA applications.

\begin{table}
\centering
\caption{Ablation Results on BAND-2k Database. The Best Results are Marked in Boldface}
\label{ablation}
\begin{tabular}{l|ccc|cc}
\toprule[0.75pt]
\multirow{2}{*}{\textsc{Model}} &\multicolumn{3}{c|}{\textbf{Patch-Level Classification}}&\multicolumn{2}{c}{\textbf{Quality Assessment}} \\ 
&AUROC &Accuracy&Speed&\makebox[0.063\textwidth][c]{SRCC}&\makebox[0.063\textwidth][c]{PLCC} \\
\midrule[0.5pt]
SB-HFM&0.8489&84.26\%&\textbf{0.0219}&0.8456&0.8301\\
SB-LFM&0.8123&82.21\%&0.0228&0.8208&0.7712\\
SB-I&0.9129&89.21\%&0.0235&0.8615&0.8656\\
DB-HFM&0.8548&84.35\%&0.0246&0.8467&0.8321\\
DB-LFM&0.8251&82.57\%&0.0271&0.8265&0.7734\\
Ours&\textbf{0.9527}&\textbf{94.18\%}&0.0252&\textbf{0.8775}&\textbf{0.8787}\\
\bottomrule[0.75pt]
\end{tabular}
\end{table}

\begin{table}
\centering
\caption{Performance Comparison of the Cross-Database Validation by Training on the Database from Kapoor \textit{et al.} \cite{kapoor2021capturing} and Testing on the BAND-2k Database.}
\label{cross1}
\begin{tabular}{l|ccc|cc}
\toprule[0.75pt]
\multirow{2}{*}{\textsc{Model}} &\multicolumn{3}{c|}{\textbf{Patch-Level Classification}}&\multicolumn{2}{c}{\textbf{Quality Assessment}} \\ 
&AUROC &Accuracy&Speed&\makebox[0.063\textwidth][c]{SRCC}&\makebox[0.063\textwidth][c]{PLCC} \\
\midrule[0.5pt]
DBI \cite{kapoor2021capturing} &0.9011&88.23\%&\textbf{0.0231}&0.7427&0.7444\\
Ours&\textbf{0.9463}&\textbf{93.39\%}&0.0253&\textbf{0.8735}&\textbf{0.8712}\\
\bottomrule[0.75pt]
\end{tabular}
\end{table}

\begin{table}[!t]
\centering
\caption{Performance Comparison of the Cross-Database Validation by Training on the BAND-2k Database and Testing on the Database from Kapoor \textit{et al.} \cite{kapoor2021capturing}}
\label{cross2}
\renewcommand\arraystretch{0.9}
\begin{tabular}{l|cccc}
\toprule[0.75pt]
\multirow{2}{*}{\textsc{Model}} &\multicolumn{4}{c}{\textbf{Patch-Level Classification}} \\ 
&AUROC &AUPRC &Accuracy &Speed \\
\midrule[0.5pt]
DBCNN \cite{zhang2018blind}&0.7481&0.7366&75.02\%&\textbf{0.0149}\\
HyperIQA \cite{su2020blindly}&0.7657&0.7634&79.33\%&0.1756\\
StairIQA \cite{sun2023blind} &0.7181&0.7018&68.62\%&0.1053\\
DBI \cite{kapoor2021capturing} &0.9462&0.9432&91.52\%&0.0233\\
Ours&\textbf{0.9886}&\textbf{0.9847}&\textbf{96.52\%}&0.0252\\
\bottomrule[0.75pt]
\end{tabular}
\end{table}

\begin{table*}
\centering
\caption{Comparison of Computational Complexity for FR and NR IQA Methods on BAND-2k Database. Time: Seconds/Image}
\label{time}
\renewcommand\arraystretch{0.8}
\begin{tabular}{lc|cccc|cccccc|ccccc}
\toprule[0.75pt]
\multicolumn{2}{c|}{\multirow{2}{*}{\scriptsize \textsc{Method}}}&\multicolumn{4}{c|}{\scriptsize \textbf{Full Reference}}&\multicolumn{6}{c|}{\scriptsize \textbf{No Reference}}&\multicolumn{5}{c}{\scriptsize \textbf{Banding-Specified}} \\ 
\multicolumn{2}{c|}{} &\tiny PSNR&\tiny SSIM&\tiny MS-SSIM&\tiny LPIPS&\tiny BRISQUE&\tiny NIQE&\tiny NIMA&\tiny DBCNN&\tiny HyberIQA&\tiny StairIQA&\tiny BBAND&\tiny CAMBI&\tiny VMAF\textsubscript{BA}&\tiny DBI&\tiny Ours\\
\midrule[0.5pt]
\multicolumn{2}{c|}{\scriptsize Time}&\scriptsize 0.1501&\scriptsize 0.1753&\scriptsize 0.2112&\scriptsize 
0.1727&\scriptsize 0.1183&\scriptsize 0.2252&\scriptsize 0.3303&\scriptsize 0.3516&\scriptsize 0.6986&\scriptsize 0.5783&\scriptsize 1.0321&\scriptsize 0.0942&\scriptsize 0.1951&\scriptsize 26.6553&\scriptsize 3.3422\\
\bottomrule[0.75pt]
\end{tabular}
\end{table*}

\subsection{Cross-Database Validation}
Due to the effects of different compression techniques, shooting equipment, scenes, etc., the image content and banding distortions may vary significantly in practical applications. For the database \cite{kapoor2021capturing}, it only includes limited types of image sources and means of triggering banding distortion. As a result, we conduct a cross-database validation to verify the generalizability of the proposed model, wherein the database presented by \cite{kapoor2021capturing} and BAND-2k are included. That is, we trained the model on one full database and report the test performance on the other. We mainly compare the proposed method with four learning-based models, i.e., DBCNN \cite{zhang2018blind},  HyperIQA \cite{su2020blindly}, StairIQA \cite{sun2023blind}, and DBI \cite{kapoor2021capturing}. Since MOS information is not provided in the database \cite{kapoor2021capturing}, which is an essential part of methods training, we condensed part of the experiments. Table \ref{cross1} and \ref{cross2} report the experimental results in terms of patch-level banding classification and image quality assessment.
We can observe that our proposed method generalization between database \cite{kapoor2021capturing} and BAND-2k was surprisingly good. Besides, it is worth noting that the performance of these methods trained on the BAND-2k has improved a little compared to the previous versions that were trained on the database \cite{kapoor2021capturing}, which further demonstrates the superiority of the proposed database BAND-2k.


\subsection{Computational Complexity}
The efficiency of an image quality prediction model is of great importance in practical industrial deployments. Therefore, we measured the average running time of the compared IQA models, as shown in Table \ref{time}. The experiments were performed in MATLAB R2021a and Python 3.7 under Windows 10 64-bit system on a Lenovo laptop with Intel Core i5-9300HF CPU@2.4GHz, 16GB RAM, and NVIDIA GTX 1660Ti 6G GPU. It can be observed that the proposed method achieves a reasonable running time among the FR, NR, and other banding-specified IQA algorithms. Generally, the execution time of classical IQA algorithms is significantly less than learning-based methods. Simpler NSS-based models such as BRISQUE and NIQE still show competitive efficiency relative to CNN models while exhibiting inferior performance in banding image quality assessment. For CAMBI and VMAF\textsubscript{BA}, we use the officially launched software package, which is based on the stand-alone C library \textit{libvmaf} and therefore surpasses other methods in speed.
Moreover, unlike the general quality evaluation using regression to predict scores, the patch-wise prediction strategy that we adopted may increase the complexity. Note that although we deployed a more complex network structure, a nearly 10 times speedup has been seen when comparing DBI with our method since the sliding window mechanism \cite{kapoor2021capturing} is removed.

\section{Conclusion}
In this paper, we conduct a comprehensive exploration of banding images from both subjective and objective perspectives. Specifically, we construct the largest ecologically valid banding IQA database to date named BAND-2k database, which consists of 2,000 banding images generated by fifteen compression and quantization schemes, achieving several times larger in number and diversity than the existing banding dataset. The construction process of the database, including distortion content preparation, subjective test procedure, and the removal of outlying data, is described in detail in this paper. Relying on this database, we proposed a novel banding evaluator using the frequency characteristic of banding artifacts, which models the banding as high-frequency artifacts that contained in the low-frequency smoothing region. A dual-branch CNN is devised to extract hierarchical features to classify the banding regions, upon which we introduce the spatial frequency masking to refine and compute an overall banding score.
Experimental results show that our proposed method outperforms the baseline algorithms significantly in patch-level banding classification and banding IQA tasks. We believe that our study will benefit further development, calibration, and benchmarking of banding IQA models.





\bibliographystyle{IEEEtran}

\begin{thebibliography}{10}
\providecommand{\url}[1]{#1}
\csname url@samestyle\endcsname
\providecommand{\newblock}{\relax}
\providecommand{\bibinfo}[2]{#2}
\providecommand{\BIBentrySTDinterwordspacing}{\spaceskip=0pt\relax}
\providecommand{\BIBentryALTinterwordstretchfactor}{4}
\providecommand{\BIBentryALTinterwordspacing}{\spaceskip=\fontdimen2\font plus
\BIBentryALTinterwordstretchfactor\fontdimen3\font minus \fontdimen4\font\relax}
\providecommand{\BIBforeignlanguage}[2]{{%
\expandafter\ifx\csname l@#1\endcsname\relax
\typeout{** WARNING: IEEEtran.bst: No hyphenation pattern has been}%
\typeout{** loaded for the language `#1'. Using the pattern for}%
\typeout{** the default language instead.}%
\else
\language=\csname l@#1\endcsname
\fi
#2}}
\providecommand{\BIBdecl}{\relax}
\BIBdecl

\bibitem{sun2023blind}
W.~Sun, X.~Min, D.~Tu, S.~Ma, and G.~Zhai, ``Blind quality assessment for in-the-wild images via hierarchical feature fusion and iterative mixed database training,'' \emph{IEEE Journal of Selected Topics in Signal Processing}, 2023.

\bibitem{yang2021full}
J.~Yang, Z.~Bian, Y.~Zhao, W.~Lu, and X.~Gao, ``Full-reference quality assessment for screen content images based on the concept of global-guidance and local-adjustment,'' \emph{IEEE Transactions on Broadcasting}, vol.~67, no.~3, pp. 696--709, 2021.

\bibitem{ni2018gabor}
Z.~Ni, H.~Zeng, L.~Ma, J.~Hou, J.~Chen, and K.-K. Ma, ``A gabor feature-based quality assessment model for the screen content images,'' \emph{IEEE Transactions on Image Processing}, vol.~27, no.~9, pp. 4516--4528, 2018.

\bibitem{ponomarenko2015image}
N.~Ponomarenko, L.~Jin, O.~Ieremeiev, V.~Lukin, K.~Egiazarian, J.~Astola, B.~Vozel, K.~Chehdi, M.~Carli, F.~Battisti \emph{et~al.}, ``Image database tid2013: Peculiarities, results and perspectives,'' \emph{Signal processing: Image communication}, vol.~30, pp. 57--77, 2015.

\bibitem{ciancio2010no}
A.~Ciancio, E.~A. da~Silva, A.~Said, R.~Samadani, P.~Obrador \emph{et~al.}, ``No-reference blur assessment of digital pictures based on multifeature classifiers,'' \emph{IEEE Transactions on image processing}, vol.~20, no.~1, pp. 64--75, 2010.

\bibitem{ghadiyaram2015massive}
D.~Ghadiyaram and A.~C. Bovik, ``Massive online crowdsourced study of subjective and objective picture quality,'' \emph{IEEE Transactions on Image Processing}, vol.~25, no.~1, pp. 372--387, 2015.

\bibitem{seshadrinathan2010study}
K.~Seshadrinathan, R.~Soundararajan, A.~C. Bovik, and L.~K. Cormack, ``Study of subjective and objective quality assessment of video,'' \emph{IEEE transactions on Image Processing}, vol.~19, no.~6, pp. 1427--1441, 2010.

\bibitem{talebi2018nima}
H.~Talebi and P.~Milanfar, ``Nima: Neural image assessment,'' \emph{IEEE transactions on image processing}, vol.~27, no.~8, pp. 3998--4011, 2018.

\bibitem{su2020blindly}
S.~Su, Q.~Yan, Y.~Zhu, C.~Zhang, X.~Ge, J.~Sun, and Y.~Zhang, ``Blindly assess image quality in the wild guided by a self-adaptive hyper network,'' in \emph{Proceedings of the IEEE/CVF Conference on Computer Vision and Pattern Recognition}, 2020, pp. 3667--3676.

\bibitem{zhang2018unreasonable}
R.~Zhang, P.~Isola, A.~A. Efros, E.~Shechtman, and O.~Wang, ``The unreasonable effectiveness of deep features as a perceptual metric,'' in \emph{Proceedings of the IEEE conference on computer vision and pattern recognition}, 2018, pp. 586--595.

\bibitem{wang2003multiscale}
Z.~Wang, E.~P. Simoncelli, and A.~C. Bovik, ``Multiscale structural similarity for image quality assessment,'' in \emph{The Thrity-Seventh Asilomar Conference on Signals, Systems \& Computers, 2003}, vol.~2.\hskip 1em plus 0.5em minus 0.4em\relax Ieee, 2003, pp. 1398--1402.

\bibitem{wang2004image}
Z.~Wang, A.~C. Bovik, H.~R. Sheikh, and E.~P. Simoncelli, ``Image quality assessment: from error visibility to structural similarity,'' \emph{IEEE transactions on image processing}, vol.~13, no.~4, pp. 600--612, 2004.

\bibitem{paszke2017automatic}
A.~Paszke, S.~Gross, S.~Chintala, G.~Chanan, E.~Yang, Z.~DeVito, Z.~Lin, A.~Desmaison, L.~Antiga, and A.~Lerer, ``Automatic differentiation in pytorch,'' 2017.

\bibitem{ghadiyaram2017no}
D.~Ghadiyaram, C.~Chen, S.~Inguva, and A.~Kokaram, ``A no-reference video quality predictor for compression and scaling artifacts,'' in \emph{2017 IEEE International Conference on Image Processing (ICIP)}.\hskip 1em plus 0.5em minus 0.4em\relax IEEE, 2017, pp. 3445--3449.

\bibitem{kazemi2022multifocus}
V.~Kazemi, A.~Shahzadi, and H.~K. Bizaki, ``Multifocus image fusion using adaptive block compressive sensing by combining spatial frequency,'' \emph{Multimedia Tools and Applications}, vol.~81, no.~11, pp. 15\,153--15\,170, 2022.

\bibitem{li2008multifocus}
S.~Li and B.~Yang, ``Multifocus image fusion using region segmentation and spatial frequency,'' \emph{Image and vision computing}, vol.~26, no.~7, pp. 971--979, 2008.

\bibitem{zhang2022dual}
C.~Zhang, Z.~Huang, S.~Liu, and J.~Xiao, ``Dual-channel multi-task cnn for no-reference screen content image quality assessment,'' \emph{IEEE Transactions on Circuits and Systems for Video Technology}, vol.~32, no.~8, pp. 5011--5025, 2022.

\bibitem{he2016deep}
K.~He, X.~Zhang, S.~Ren, and J.~Sun, ``Deep residual learning for image recognition,'' in \emph{Proceedings of the IEEE conference on computer vision and pattern recognition}, 2016, pp. 770--778.

\bibitem{bar2006semi}
L.~Bar, N.~Sochen, and N.~Kiryati, ``Semi-blind image restoration via mumford-shah regularization,'' \emph{IEEE Transactions on Image Processing}, vol.~15, no.~2, pp. 483--493, 2006.

\bibitem{al2017multi}
S.~A. Al-Sumaidaee, M.~A. Abdullah, R.~R.~O. Al-Nima, S.~S. Dlay, and J.~A. Chambers, ``Multi-gradient features and elongated quinary pattern encoding for image-based facial expression recognition,'' \emph{Pattern Recognition}, vol.~71, pp. 249--263, 2017.

\bibitem{gu2017evaluating}
K.~Gu, J.~Qiao, X.~Min, G.~Yue, W.~Lin, and D.~Thalmann, ``Evaluating quality of screen content images via structural variation analysis,'' \emph{IEEE transactions on visualization and computer graphics}, vol.~24, no.~10, pp. 2689--2701, 2017.

\bibitem{tang2018full}
Z.~Tang, Y.~Zheng, K.~Gu, K.~Liao, W.~Wang, and M.~Yu, ``Full-reference image quality assessment by combining features in spatial and frequency domains,'' \emph{IEEE Transactions on Broadcasting}, vol.~65, no.~1, pp. 138--151, 2018.

\bibitem{grubbs1950sample}
F.~E. Grubbs, ``Sample criteria for testing outlying observations,'' \emph{The Annals of Mathematical Statistics}, pp. 27--58, 1950.

\bibitem{narvekar2011no}
N.~D. Narvekar and L.~J. Karam, ``A no-reference image blur metric based on the cumulative probability of blur detection (cpbd),'' \emph{IEEE Transactions on Image Processing}, vol.~20, no.~9, pp. 2678--2683, 2011.

\bibitem{hasler2003measuring}
D.~Hasler and S.~E. Suesstrunk, ``Measuring colorfulness in natural images,'' in \emph{Human vision and electronic imaging VIII}, vol. 5007.\hskip 1em plus 0.5em minus 0.4em\relax SPIE, 2003, pp. 87--95.

\bibitem{bt2002methodology}
R.~I.-R. BT, ``Methodology for the subjective assessment of the quality of television pictures,'' \emph{International Telecommunication Union}, 2002.

\bibitem{sullivan2012overview}
G.~J. Sullivan, J.-R. Ohm, W.-J. Han, and T.~Wiegand, ``Overview of the high efficiency video coding (hevc) standard,'' \emph{IEEE Transactions on circuits and systems for video technology}, vol.~22, no.~12, pp. 1649--1668, 2012.

\bibitem{wiegand2003overview}
T.~Wiegand, G.~J. Sullivan, G.~Bjontegaard, and A.~Luthra, ``Overview of the h. 264/avc video coding standard,'' \emph{IEEE Transactions on circuits and systems for video technology}, vol.~13, no.~7, pp. 560--576, 2003.

\bibitem{hosu2020koniq}
V.~Hosu, H.~Lin, T.~Sziranyi, and D.~Saupe, ``Koniq-10k: An ecologically valid database for deep learning of blind image quality assessment,'' \emph{IEEE Transactions on Image Processing}, vol.~29, pp. 4041--4056, 2020.

\bibitem{bhagavathy2009multiscale}
S.~Bhagavathy, J.~Llach, and J.~Zhai, ``Multiscale probabilistic dithering for suppressing contour artifacts in digital images,'' \emph{IEEE Transactions on Image Processing}, vol.~18, no.~9, pp. 1936--1945, 2009.

\bibitem{huang2016understanding}
Q.~Huang, H.~Y. Kim, W.-J. Tsai, S.~Y. Jeong, J.~S. Choi, and C.-C.~J. Kuo, ``Understanding and removal of false contour in hevc compressed images,'' \emph{IEEE Transactions on Circuits and Systems for Video Technology}, vol.~28, no.~2, pp. 378--391, 2016.

\bibitem{lee2006two}
J.~W. Lee, B.~R. Lim, R.-H. Park, J.-S. Kim, and W.~Ahn, ``Two-stage false contour detection using directional contrast and its application to adaptive false contour reduction,'' \emph{IEEE Transactions on Consumer Electronics}, vol.~52, no.~1, pp. 179--188, 2006.

\bibitem{daly2004decontouring}
S.~J. Daly and X.~Feng, ``Decontouring: Prevention and removal of false contour artifacts,'' in \emph{Human Vision and Electronic Imaging IX}, vol. 5292.\hskip 1em plus 0.5em minus 0.4em\relax SPIE, 2004, pp. 130--149.

\bibitem{baugh2014advanced}
G.~Baugh, A.~Kokaram, and F.~Piti{\'e}, ``Advanced video debanding,'' in \emph{Proceedings of the 11th European Conference on Visual Media Production}, 2014, pp. 1--10.

\bibitem{zhang2018blind}
W.~Zhang, K.~Ma, J.~Yan, D.~Deng, and Z.~Wang, ``Blind image quality assessment using a deep bilinear convolutional neural network,'' \emph{IEEE Transactions on Circuits and Systems for Video Technology}, vol.~30, no.~1, pp. 36--47, 2018.

\bibitem{mittal2012making}
A.~Mittal, R.~Soundararajan, and A.~C. Bovik, ``Making a ``completely blind'' image quality analyzer,'' \emph{IEEE Signal processing letters}, vol.~20, no.~3, pp. 209--212, 2012.

\bibitem{mittal2012no}
A.~Mittal, A.~K. Moorthy, and A.~C. Bovik, ``No-reference image quality assessment in the spatial domain,'' \emph{IEEE Transactions on image processing}, vol.~21, no.~12, pp. 4695--4708, 2012.

\bibitem{mukherjee2013latest}
D.~Mukherjee, J.~Bankoski, A.~Grange, J.~Han, J.~Koleszar, P.~Wilkins, Y.~Xu, and R.~Bultje, ``The latest open-source video codec vp9-an overview and preliminary results,'' in \emph{2013 Picture Coding Symposium (PCS)}.\hskip 1em plus 0.5em minus 0.4em\relax IEEE, 2013, pp. 390--393.

\bibitem{wang2016perceptual}
Y.~Wang, S.-U. Kum, C.~Chen, and A.~Kokaram, ``A perceptual visibility metric for banding artifacts,'' in \emph{2016 IEEE International Conference on Image Processing (ICIP)}.\hskip 1em plus 0.5em minus 0.4em\relax IEEE, 2016, pp. 2067--2071.

\bibitem{krasula2022banding}
L.~Krasula, Z.~Li, C.~G. Bampis, M.~Afonso, N.~F. Miret, and J.~Sole, ``Banding vs. quality: perceptual impact and objective assessment,'' in \emph{2022 IEEE International Conference on Image Processing (ICIP)}.\hskip 1em plus 0.5em minus 0.4em\relax IEEE, 2022, pp. 2236--2240.

\bibitem{tandon2021cambi}
P.~Tandon, M.~Afonso, J.~Sole, and L.~Krasula, ``Cambi: Contrast-aware multiscale banding index,'' in \emph{2021 Picture Coding Symposium (PCS)}.\hskip 1em plus 0.5em minus 0.4em\relax IEEE, 2021, pp. 1--5.

\bibitem{tu2020bband}
Z.~Tu, J.~Lin, Y.~Wang, B.~Adsumilli, and A.~C. Bovik, ``Bband index: A no-reference banding artifact predictor,'' in \emph{ICASSP 2020-2020 IEEE International Conference on Acoustics, Speech and Signal Processing (ICASSP)}.\hskip 1em plus 0.5em minus 0.4em\relax IEEE, 2020, pp. 2712--2716.

\bibitem{kapoor2021capturing}
A.~Kapoor, J.~Sapra, and Z.~Wang, ``Capturing banding in images: Database construction and objective assessment,'' in \emph{ICASSP 2021-2021 IEEE International Conference on Acoustics, Speech and Signal Processing (ICASSP)}.\hskip 1em plus 0.5em minus 0.4em\relax IEEE, 2021, pp. 2425--2429.

\bibitem{deyoe1988concurrent}
E.~A. DeYoe and D.~C. Van~Essen, ``Concurrent processing streams in monkey visual cortex,'' \emph{Trends in neurosciences}, vol.~11, no.~5, pp. 219--226, 1988.

\bibitem{wang2017videoset}
H.~Wang, I.~Katsavounidis, J.~Zhou, J.~Park, S.~Lei, X.~Zhou, M.-O. Pun, X.~Jin, R.~Wang, X.~Wang \emph{et~al.}, ``Videoset: A large-scale compressed video quality dataset based on jnd measurement,'' \emph{Journal of Visual Communication and Image Representation}, vol.~46, pp. 292--302, 2017.

\bibitem{ranjan2017hyperface}
R.~Ranjan, V.~M. Patel, and R.~Chellappa, ``Hyperface: A deep multi-task learning framework for face detection, landmark localization, pose estimation, and gender recognition,'' \emph{IEEE transactions on pattern analysis and machine intelligence}, vol.~41, no.~1, pp. 121--135, 2017.

\bibitem{zeiler2014visualizing}
M.~D. Zeiler and R.~Fergus, ``Visualizing and understanding convolutional networks,'' in \emph{Computer Vision--ECCV 2014: 13th European Conference, Zurich, Switzerland, September 6-12, 2014, Proceedings, Part I 13}.\hskip 1em plus 0.5em minus 0.4em\relax Springer, 2014, pp. 818--833.

\bibitem{bentley1975multidimensional}
J.~L. Bentley, ``Multidimensional binary search trees used for associative searching,'' \emph{Communications of the ACM}, vol.~18, no.~9, pp. 509--517, 1975.

\end{thebibliography}


\end{document}